\documentclass[10pt,twocolumn,letterpaper]{article}

\usepackage[pagenumbers]{cvpr} 

\usepackage{colortbl}   
\usepackage{array}
\usepackage{tabularx}
\usepackage{multirow}
\usepackage{makecell}
\usepackage{comment}
\usepackage{adjustbox}
\usepackage{pgfplots}
\usepackage{wrapfig}
\usepgfplotslibrary{groupplots}
\usepackage{fontawesome5}
\pgfplotsset{compat=1.18}

\newif\ifshownotes
\shownotestrue
\newcounter{notecount}

\definecolor{ErenCol}{named}{RoyalBlue}
\definecolor{FangzhouCol}{named}{ForestGreen}
\definecolor{IvoCol}{named}{RedOrange}

\definecolor{RemiCol}{named}{Plum}
\definecolor{MahdiCol}{named}{TealBlue}
\definecolor{ChiaraCol}{named}{Magenta}
\definecolor{GabrieleCol}{named}{Brown}

\AtEndDocument{%
  \ifshownotes
    \typeout{^^J*** WARNING: \thenotecount\space annotations are VISIBLE.
      Set \string\shownotesfalse\space before submitting. ***^^J}%
  \fi}

\definecolor{visualgrounding}{RGB}{105,130,145}
\definecolor{temporalgrounding}{RGB}{151,130,75}
\definecolor{scenelocalization}{RGB}{105,135,112}
\definecolor{spatialperception}{RGB}{105,88,125}

\newcommand{\qnum}[2]{%
  \tikz[baseline=(num.base)]{
    \node[
      circle, fill=#1, text=white,
      font=\bfseries\scriptsize,
      inner sep=1.2pt, minimum size=1em
    ] (num) {#2};
  }%
}

\newcommand{\capability}[2]{%
  \par\smallskip\noindent
  \begingroup
  \setlength{\fboxsep}{2.5pt}%
  \colorbox{#1!18}{%
    \textcolor{#1!75!black}{\textbf{\strut #2.}}%
  }%
  \endgroup
  \hspace{0.3em}%
}

\definecolor{cvprblue}{rgb}{0.21,0.49,0.74}
\usepackage[pagebackref,breaklinks,colorlinks,allcolors=cvprblue]{hyperref}

\def\paperID{239} 
\def\confName{3DV\xspace}
\def\confYear{2027\xspace}

\title{Long Time No See: Benchmarking VLMs for Out-of-Sight Spatiotemporal Reasoning in Egocentric Videos}

\author{
Fangzhou Ma$^{*1}$ \quad
Ivo Alexander Ban$^{*1}$ \quad
Eren Homburg$^{*1}$ \quad
Gabriele Goletto$^{2}$ \quad
R\'emi Pautrat$^{2}$ \\[0.3em]
Mahdi Rad$^{2}$ \quad
Chiara Plizzari$^{3}$ \quad
Marc Pollefeys$^{1,2}$ \\[0.5em]
$^{1}$ETH Zurich \quad
$^{2}$Microsoft Spatial AI Lab \quad
$^{3}$Bocconi University \\[0.3em]
{\tt\small \{fangma, ivoban, ehomburg\}@student.ethz.ch}
}

\begin{document}
\makeatletter
\twocolumn[{%
    \@maketitle

    \vspace{-2em}

    {\centering
    \small
    \href{https://beyond3d-bench.github.io/website/}{\faGlobe\ Project Page}
    \quad$\cdot$\quad
    \href{https://github.com/Beyond3D-bench/vlm-evaluation}{\faGithub\ Code}
    \quad$\cdot$\quad
    \href{https://huggingface.co/datasets/Ffffangzhu/BEYOND3D}{\faDatabase\ \textsc{Beyond3D} Benchmark}
    \par}

    \vspace{1em}

    \centering
    
    \includegraphics[width=0.95\textwidth]{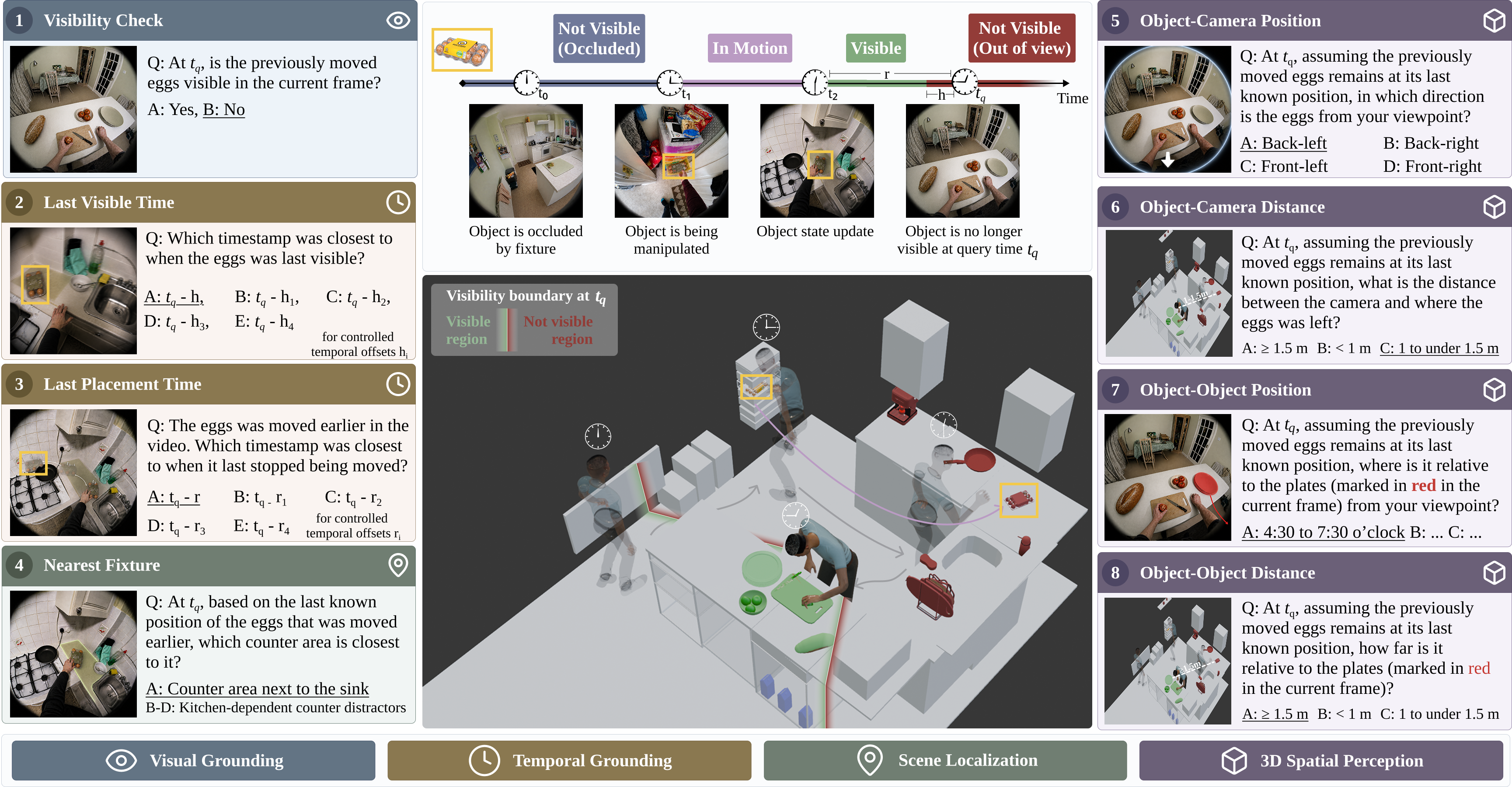}

    \captionsetup{hypcap=false}
    \captionof{figure}{\textbf{Visual illustration of the \textsc{Beyond3D} benchmark.}
    We propose a benchmark that evaluates whether VLMs can follow interacted objects and reason about their location once they leave view. As an object (here, the box of eggs) is moved by the person, the benchmark poses eight questions that progressively probe the reasoning needed to recover its state once it gets out of sight. Answering them requires VLMs to \emph{track} actively manipulated objects, \emph{update} their spatial state after relocation, and \emph{recall} that state once the objects leave view.}
    \captionsetup{hypcap=true}
    \label{fig:pipeline}

    \vspace{0.7em}
}]
\makeatother

\iftoggle{cvprfinal}{%
  \renewcommand{\thefootnote}{\fnsymbol{footnote}}%
  \footnotetext[1]{Equal contribution.}%
  \renewcommand{\thefootnote}{\arabic{footnote}}%
}{}

\begin{abstract}
Real-world AI systems must reason about objects that are no longer visible: an AR assistant guiding a user back to an object used earlier, a household robot retrieving an item someone put away. This requires not just recalling where an object was last seen, but updating its state when it is moved and retaining that update once it leaves view. We refer to this as \textbf{out-of-sight spatiotemporal reasoning}. We introduce \textsc{Beyond3D}, the first VQA benchmark to isolate this ability in dynamic egocentric video: every query targets an object that has been relocated and has since left the field of view. We create our questions from HD-EPIC annotations, building a visibility track for each dynamic object from its 3D position, the camera pose, and the scene geometry to understand at each moment whether it is visible, occluded, or out of view. \textsc{Beyond3D} comprises 9,000 questions in eight types over 135 videos from nine participants, organized as one reasoning chain: visual grounding (is the target observable now), temporal grounding (when it was last visible and last placed), scene localization (which fixture anchors that location), and 3D spatial perception (where it lies relative to the current viewpoint or another object in the scene). We benchmark nine general-purpose and spatially specialized VLMs. The best model reaches 42.2\% against 29.7\% chance and text-only baselines reaching 31.9\%, with the largest failures in recovering when an object was last visible, showing that tracking object movement out of sight remains far from solved for current VLMs.

\end{abstract}
\section{Introduction}
\label{sec:intro}
For an embodied system, understanding only what is currently visible is not enough. Humans can remember and reason about previously seen objects even after they leave sight~\cite{burgess2006spatial,plizzari2025osnom}. Similarly, an AR assistant may need to guide a user back to an object handled earlier, while a household robot may need to retrieve an item after it has been moved out of sight. In such cases, the system must identify the interactions that established the object’s latest location and retain that state as the scene evolves. This is particularly challenging in egocentric video, where objects are manipulated and relocated as the viewpoint continuously changes. We refer to this ability to reason about the evolving spatial state of objects beyond the current field of view as \textbf{out-of-sight spatiotemporal reasoning}.

Recent vision-language models (VLMs) are increasingly capable of recognizing, describing, and answering questions about visual content~\cite{qwen2026qwen35,qwen2026qwen36}. However, it remains unclear whether they can maintain coherent spatial representations of dynamic environments and reason about out-of-sight objects. Existing benchmarks cover related aspects of memory, temporal reasoning, and 3D scene understanding, but do not directly test if models retain the updated spatial state of a relocated object after it leaves view (Table~\ref{tab:benchmark_comparison}).

We introduce \textsc{Beyond3D}, a Visual Question Answering (VQA) benchmark designed to isolate out-of-sight spatiotemporal reasoning in dynamic egocentric video. We build on HD-EPIC~\cite{perrett2025hdepic}, which captures unscripted cooking and everyday activities with detailed annotations of object movements and 3D scene reconstructions. From these recordings, we identify interactions in which objects are picked up, carried, and relocated across counters, cupboards, drawers, etc. For each relocated object, we construct a \textit{geometry-aware visibility track} combining its 3D location with the camera pose, field of view, and scene geometry to determine whether it is \textit{visible}, \textit{occluded}, or \textit{outside the camera view} at each time step. These tracks allow us to follow the object’s spatial state as the wearer continues interacting with the environment and, crucially, to place queries only after the object is no longer observable.

\textsc{Beyond3D} comprises eight question types covering \emph{visual grounding}, \emph{temporal grounding}, \emph{scene localization}, and \emph{3D spatial perception}. Together, they trace the reasoning chain in Fig.~\ref{fig:pipeline}, from identifying the relevant event and recovering the object’s last location to reasoning about its spatial relationships after it leaves the field of view.

Our contributions are threefold:
\begin{itemize}
\item \textbf{A new problem.} We formulate \emph{out-of-sight spatiotemporal reasoning}: tracking the spatial state of objects after they leave the field of view, a fundamental capability for reasoning in dynamic, embodied environments.

\item \textbf{A large-scale benchmark.} We introduce \textsc{Beyond3D}, comprising 9,000 questions on visual and temporal grounding, scene localization, and 3D spatial perception. Geometry-aware visibility tracks combine object locations, camera poses, fields of view, and occlusion reasoning to determine object visibility over time.

\item \textbf{A systematic evaluation of current VLMs.} We benchmark nine general-purpose and spatially specialized VLMs and analyze their failure modes. Temporal retrieval and spatial-state maintenance remain major bottlenecks, while substantial 3D reasoning errors persist even when upstream uncertainty is controlled.
\end{itemize}

\section{Related Work}
\label{sec:related}

\par\noindent
\textbf{Persistent dynamic world modeling in egocentric perception.}
Operating in dynamic environments requires reasoning beyond what is currently visible. In cognitive science, this relates to \emph{object permanence}~\cite{baillargeon1986representing}: understanding that objects continue to exist when unseen. In machine perception, it means maintaining object representations after they leave the field of view~\cite{tokmakov2022object}. In interactive environments, however, persistence alone is insufficient because manipulation and relocation can change object state. The Event Calculus~\cite{kowalski1986logic} captures this by treating world state as persistent until an event changes it. A coherent visual world model must therefore retain information about unseen objects and update their state when interactions alter it.

Egocentric video makes this particularly challenging: wearer motion changes visibility while interactions change object locations. Recent methods increasingly address this by persistent world-state representations. AMEGO~\cite{goletto2024amego} stores past interactions and visited locations in queryable memory. OSNOM~\cite{plizzari2025osnom} maintains persistent 3D locations of active objects, while Whareformer~\cite{chalk2026whareformer} replaces this engineered state maintenance with learned updates to persistent object representations and 3D positions over time.

\par\noindent
\textbf{Vision language models (VLMs).}
Recent general-purpose VLMs have expanded toward longer video inputs and stronger temporal modeling~\cite{bai2025qwen25vl,bai2025qwen3vl,qwen2026qwen35,qwen2026qwen36,wang2025internvl35,li2024llavaonevision,clark2026molmo2}, but explicit 3D spatial modeling is typically not their primary objective. Spatially specialized VLMs introduce spatial structure through targeted supervision~\cite{chen2024spatialvlm,cai2025sensenovasi,yang2026cambrians,yang2026cambrianp}, learned geometry representations~\cite{wu2025spatialmllm,zheng2025learning,fan2025vlm3r,yu2026stream3dvlm,zhao2026spacemind,qu2026loc3rvlm,gu2026spacemind,gwak2026cog3dmap}, or explicit 3D inputs such as depth and camera pose~\cite{cheng2024spatialrgpt,chen2025sdvlm,lin2026qwen3d}.

\par\noindent
\textbf{Comparison with existing benchmarks.}
Related benchmarks differ in the states that models must recover at query time (Table~\ref{tab:benchmark_comparison}). \textbf{Temporal and object memory benchmarks} test event timing or past observations without grounding memory in 3D~\cite{plizzari2025egotempo,wang2026egomemreason}, while \textbf{static 3D benchmarks} focus on scene geometry and layout~\cite{yang2025vsibench}. \textbf{Dynamic spatial-state benchmarks} capture changing configurations, including object relocations~\cite{yuan2025eocbench,perrett2025hdepic,huang2026egodynamic4d,wang2026ucsbench,zhao2026spatial} and user-centric relation changes~\cite{wang2026ucsbench}, but do not enforce target invisibility, allowing current-frame shortcuts. SCP-Bench~\cite{zhao2026spatial} instead infers unseen past or future states from partial video. Others also consider out-of-sight reasoning: Ego4D-VQ3D~\cite{grauman2022ego4d}, which retrieves previously observed stationary object locations without tracking updates, and SpaMEM~\cite{liao2026spamem}, which evaluates spatial-state revision in synthetic environments. Our benchmark \textsc{Beyond3D} requires both relocation and loss of visibility in unscripted real-world video, with visibility verified geometrically.

\newcommand{\cmark}{\textcolor{green!50!black}{\checkmark}}
\newcommand{\pmark}{\textcolor{orange!85!black}{$\circ$}}
\newcommand{\nmark}{\textcolor{red!65!black}{--}}
\begin{table}[t]
\centering
\caption{\textbf{Comparison with related spatial and memory benchmarks.}
\cmark: explicitly evaluated;
\pmark: partially covered or not systematically enforced;
\nmark: not explicitly targeted.
\emph{Ego.}: egocentric input;
\emph{3D}: explicit 3D spatial reasoning;
\emph{Tem-Loc.}: temporal localization;
\emph{Spa-Upd.}: spatial-state update;
\emph{Diag.}: multi-level diagnostic decomposition;
\emph{Upd-OOS.}: updated target queried after loss of visibility;
\emph{Geo-Vis.}: geometry-aware visibility.}
\label{tab:benchmark_comparison}

\scriptsize
\setlength{\tabcolsep}{1.35pt}
\renewcommand{\arraystretch}{1.08}

\begin{tabular}{
    @{}
    >{\raggedright\arraybackslash}p{2.1cm}
    !{\color{black!35}\vrule width 0.45pt}
    *{7}{>{\centering\arraybackslash}p{0.75cm}}
    @{}
}
\toprule

\textbf{Benchmark}
& \textbf{Ego.}
& \textbf{3D}
& \makecell{\textbf{Tem-}\\\textbf{Loc.}}
& \makecell{\textbf{Spa-}\\\textbf{Upd.}}
& \textbf{Diag.}
& \makecell{\textbf{Upd-}\\\textbf{OOS}}
& \makecell{\textbf{Geo-}\\\textbf{Vis.}}
\\
\midrule

\rowcolor{blue!4}
\multicolumn{8}{@{}l@{}}{
    \hspace{0.25em}\textit{(A) Temporal and object memory}
}
\\[1pt]

EgoTempo~\cite{plizzari2025egotempo}
& \cmark
& \nmark
& \pmark
& \pmark
& \nmark
& \nmark
& \nmark
\\

EgoMemReason~\cite{wang2026egomemreason}
& \cmark
& \nmark
& \pmark
& \pmark
& \nmark
& \nmark
& \nmark
\\

\addlinespace[1pt]
\midrule
\addlinespace[0.5pt]

\rowcolor{blue!4}
\multicolumn{8}{@{}l@{}}{
    \hspace{0.25em}\textit{(B) Static 3D spatial reasoning}
}
\\[1pt]

VSI-Bench~\cite{yang2025vsibench}
& \cmark
& \cmark
& \nmark
& \nmark
& \nmark
& \nmark
& \nmark
\\

\addlinespace[1pt]
\midrule
\addlinespace[0.5pt]

\rowcolor{blue!4}
\multicolumn{8}{@{}l@{}}{
    \hspace{0.25em}\textit{(C) Dynamic spatial state reasoning}
}
\\[1pt]

EOC-Bench~\cite{yuan2025eocbench}
& \cmark
& \nmark
& \cmark
& \cmark
& \nmark
& \nmark
& \nmark
\\

HD-EPIC~\cite{perrett2025hdepic}
& \cmark
& \cmark
& \cmark
& \cmark
& \nmark
& \nmark
& \nmark
\\

EgoDynamic4D~\cite{huang2026egodynamic4d}
& \cmark
& \cmark
& \pmark
& \cmark
& \nmark
& \nmark
& \nmark
\\

UCS-Bench~\cite{wang2026ucsbench}
& \cmark
& \cmark
& \pmark
& \pmark
& \pmark
& \pmark
& \nmark
\\

SCP-Bench~\cite{zhao2026spatial}
& \nmark
& \nmark
& \nmark
& \pmark
& \pmark
& \nmark
& \nmark
\\

Ego4D-VQ3D~\cite{grauman2022ego4d}
& \cmark
& \cmark
& \cmark
& \nmark
& \nmark
& \nmark
& \nmark
\\

SpaMEM~\cite{liao2026spamem}
& \nmark
& \cmark
& \cmark
& \cmark
& \cmark
& \pmark
& \pmark
\\

\addlinespace[1pt]
\midrule
\addlinespace[0.5pt]

\rowcolor{blue!8}
\textbf{\textsc{Beyond3D} (Ours)}
& \cmark
& \cmark
& \cmark
& \cmark
& \cmark
& \cmark
& \cmark
\\

\bottomrule
\end{tabular}
\end{table}
\section{Out-of-Sight Spatiotemporal Reasoning}
\label{sec:task}

\subsection{Problem Formulation}
\label{sec:problem_formulation}

\par\noindent
\textbf{Task setting.}
We consider \emph{active objects} that the camera wearer manipulates and relocates. Given an egocentric video up to $T_q$ and such a target object $o$, the task is to track its status while visible, update it upon relocation, and recall it once $o$ leaves view. 
\par\noindent
\textbf{Object state and query anchor.}
At each time $t$, the object has a 3D location $\ell_o(t)\in\mathbb{R}^3\cup\{\bot\}$, where $\bot$ marks an unannotated relocation interval and any other value means $o$ is stationary. Each stationary location is associated with a semantic fixture $f_o(t)$ that supports or contains the object, such as a counter, drawer, or appliance. 

A binary state $v_o(t)\in\{0,1\}$ records whether $o$ is visible in the frame at time $t$. An object may be not visible because it lies outside the camera view or is occluded by a hand, another object, or a cabinet door. We query objects through anchors $(o,T_q)$ for which $o$ has been relocated before $T_q$ and is stationary but not visible at query time, so that $\ell_o(T_q)\neq\bot$ and $v_o(T_q)=0$. We call $\ell_o(T_q)$ and $f_o(T_q)$ the \emph{last known} location and fixture of the target.

The out-of-sight horizon measures the time since $o$ was last visible,
\[
\begin{array}{c}
h(o,T_q)=T_q-\max\{t\leq T_q\mid v_o(t)=1\}.
\end{array}
\]
Reference-object questions use a distinct object $r\neq o$ that is stationary and visible at $T_q$, providing a scene-relative reference whose distance to $o$ is invariant to ego-motion and whose direction is invariant to ego-translation.

\subsection{\textsc{Beyond3D} Q\&A Formulation}
\label{sec:questions}
Given a query anchor $(o,T_q)$, consisting of a target object $o$ and query time $T_q$, we define eight diagnostic question types organized into four complementary capabilities: \emph{visual grounding}, \emph{temporal grounding}, \emph{scene localization}, and \emph{3D spatial perception}. As illustrated in Fig.~\ref{fig:pipeline}, these questions probe successive stages of out-of-sight reasoning: determining whether the target is currently observable, retrieving the events that established its latest status, grounding its remembered location in the scene, and reasoning about that location in 3D. The exact natural-language templates and answer choices are provided in Supp.~\ref{sec:QA_template}.

\capability{visualgrounding}{Visual Grounding}
We test whether the model recognizes the target and determines its visibility at query time:
\qnum{visualgrounding}{1}\hspace{0.1em}
\textbf{Visibility Check:} Is $o$ visible at $T_q$?

\capability{temporalgrounding}{Temporal Grounding}
We test whether the model can retrieve the events defining the target’s latest status:
\qnum{temporalgrounding}{2}\hspace{0.1em}
\textbf{Last Visible Time:} When was $o$ last visible before $T_q$?
\hspace{0.1em}
\qnum{temporalgrounding}{3}\hspace{0.1em}
\textbf{Last Placement Time:} When was $o$ last placed before $T_q$?

\capability{scenelocalization}{Scene Localization}
We test whether the model can update the target's spatial status and anchor its last known location to the scene:
\qnum{scenelocalization}{4}\hspace{0.1em}
\textbf{Nearest Fixture:} Which fixture is closest to the last known location of $o$?

\capability{spatialperception}{3D Spatial Perception}
We test whether the model can reason about the remembered target location relative to the camera and other objects:
\qnum{spatialperception}{5}\hspace{0.1em}
\textbf{Object--Camera Direction:} Where is $o$ relative to the camera at $T_q$?
\hspace{0.1em}
\qnum{spatialperception}{6}\hspace{0.1em}
\textbf{Object--Camera Distance:} How far is $o$ from the camera at $T_q$?
\hspace{0.1em}
\qnum{spatialperception}{7}\hspace{0.1em}
\textbf{Object--Object Direction:} Where is $o$ relative to reference object $r$ at $T_q$?
\hspace{0.1em}
\qnum{spatialperception}{8}\hspace{0.1em}
\textbf{Object--Object Distance:} How far is $o$ from $r$ at $T_q$?
\section{Benchmark Construction}
\label{sec:benchmark}

We build \textsc{Beyond3D} on HD-EPIC~\cite{perrett2025hdepic} in two stages as shown in Fig.~\ref{fig:Benchmark construction pipeline}: (i) infer per-object visibility tracks, (ii) select balanced out-of-sight query anchors, and finally generate questions per anchor for VLM evaluation.


\subsection{Visibility Tracks}
\label{sec:visibility}

\begin{figure}[t]
\centering
\includegraphics[width=\columnwidth]{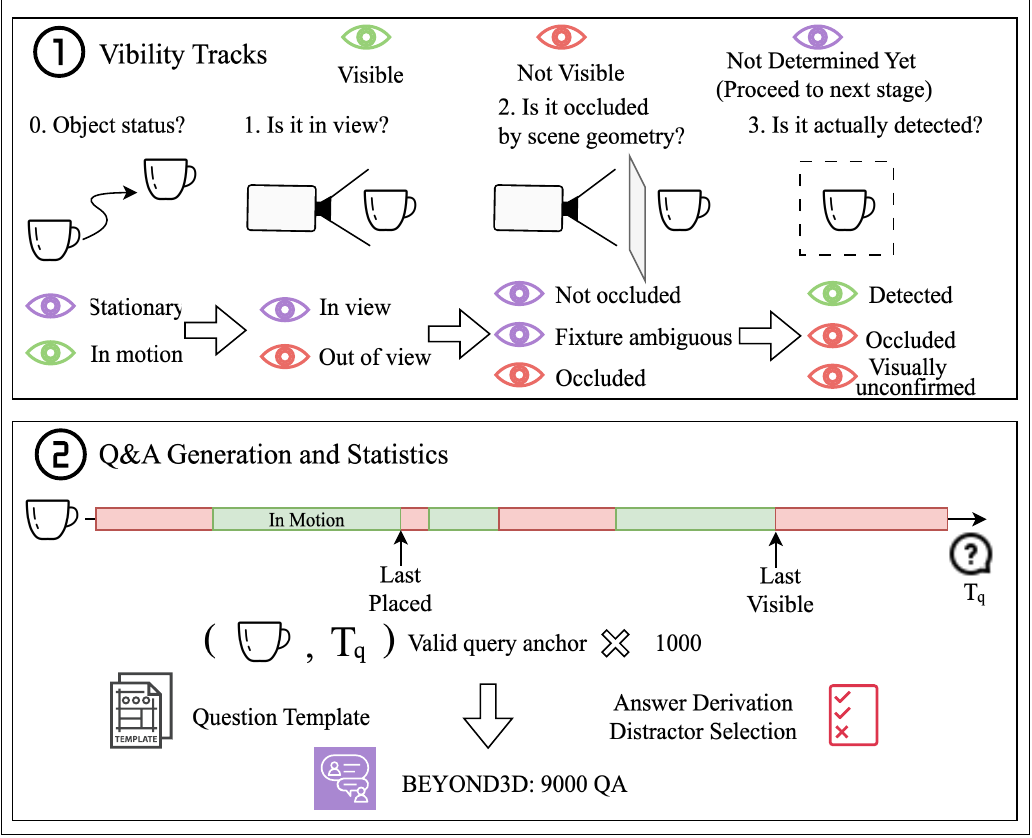}
\caption{\textbf{\textsc{Beyond3D} benchmark construction.}
\textbf{(1)} Visibility tracks are inferred through view, occlusion, and detection checks.
\textbf{(2)} Valid out-of-sight anchors are converted into 9,000 questions.}
\vspace{-15pt}
\label{fig:Benchmark construction pipeline}
\end{figure}

HD-EPIC~\cite{perrett2025hdepic} annotates object relocations in egocentric kitchen videos. For each movement, it provides the start and end times and, at both endpoints, the object's 2D bounding box, mask, 3D center, and supporting or containing \emph{fixture} (e.g., a counter or drawer). It also provides a reconstructed 3D digital twin of each kitchen. Since locations are annotated only at movement endpoints, we assume each object remains at its last annotated location until the next movement begins and mark movement intervals as \texttt{in\_motion}. Using these annotations, the digital twins, and video frames, we construct a 1\,fps \emph{visibility track} for each object $o$, recording its location $\ell_o(t)$ and visibility state $v_o(t)$ over time.

\par
\begin{figure}[t]
\centering
\includegraphics[width=\columnwidth]{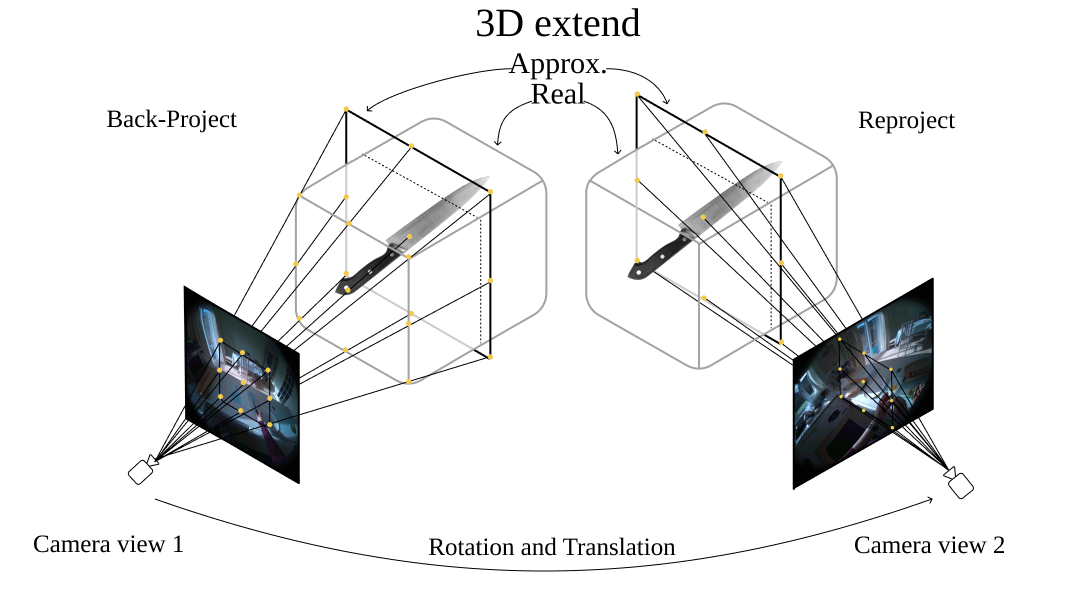}
\caption{\textbf{Cross-view object projection.}
An Knife's annotated image footprint in Camera View~1 is back-projected to its last known 3D depth and reprojected into Camera View~2 using the relative camera pose. The projected footprint approximates the object's spatial extent under the new viewpoint.}
\label{fig:triangularization}
\vspace{-15pt}
\end{figure}

\noindent
\textbf{Determining visibility.}
We determine the visibility of each stationary object in three stages (\cref{fig:Benchmark construction pipeline}):

\textbf{\emph{Stage 1: Camera field of view.}} 
We first determine whether the object projects into the current camera view. From its most recent annotated bounding box, we choose the center point, the four corners, and the four edge midpoints to approximate the object's spatial extent. These points are back-projected to the 3D scene at the depth of the object's last known position and then reprojected into the current frame using the relative camera pose and the FISHEYE624 fisheye camera model of Project Aria~\cite{engel2023projectaria}. Fig.~\ref{fig:triangularization} llustrates this back-projection and re-projection procedure
across two camera viewpoints.

Because Aria images are fisheye and vignetted, we consider only the usable circular region inscribed in the square frame. We mark an object as \texttt{out\_of\_view} when fewer than half of its projected footprint points fall inside this region. All remaining samples proceed to Stage~2.

\textbf{\emph{Stage 2: Geometric occlusion.}}
For each footprint point, we cast a ray from the camera to its corresponding 3D location and intersect it with the static kitchen mesh. As in Stage~1, we use a majority heuristic and mark the object as \texttt{occluded} when at least half of the rays are blocked. To mitigate annotation inaccuracies, we require the first intersection to lie at least $\delta=10\,\mathrm{cm}$ in front of the target.

An object may be blocked only by the fixture it currently occupies. The static meshes do not capture if fixtures like drawers or cupboards are open or closed which is why we mark these cases as \texttt{fixture\_ambiguous}. All samples which are not \texttt{occluded} proceed to Stage~3.

\textbf{\emph{Stage 3: Detection-based confirmation.}}
The static mesh does not capture transient occlusions by hands, movable objects, or clutter. We verify samples passing the geometric tests in the video using the open-vocabulary detector OWLv2~\cite{minderer2023scalingOWLv2}. Further details are provided in Supp.~\ref{sec:Supplementary_determining_visibility}.

As in the previous stages, we use a majority heuristic, marking an interval \texttt{detected\_\allowbreak visible} if the object is detected in at least half of the tested frames and \texttt{visually\_\allowbreak unconfirmed} otherwise. \texttt{fixture\_\allowbreak ambiguous} states are instead marked \texttt{occluded} after negative detection. 

\par\noindent
\textbf{Track assembly.} Consecutive samples sharing the same stage-1/stage-2 outcome are merged into intervals, which Stage~3 then labels as a whole, forming the final visibility tracks. These tracks distinguish \texttt{visible}, \texttt{out\_\allowbreak of\_\allowbreak view}, \texttt{occluded}, \texttt{visually\_\allowbreak unconfirmed}, and \texttt{in\_\allowbreak motion} states. The binary visibility of \cref{sec:problem_formulation} follows as $v_o(t)=1$ for \texttt{visible} samples and $v_o(t)=0$ otherwise.

\begin{figure*}[t!]
\centering

\definecolor{myteal}{HTML}{75C6CE}
\definecolor{myteallight}{HTML}{DFF3F5}
\definecolor{mygrid}{HTML}{ECECEC}
\definecolor{mytext}{HTML}{303030}
\definecolor{mymuted}{HTML}{777777}

\begin{adjustbox}{max width=\textwidth}
\begin{tikzpicture}

\begin{groupplot}[
    group style={
        group size=4 by 1,
        horizontal sep=1.3cm,
    },
    width=0.305\textwidth,
    height=4.4cm,
    scale only axis,
    ymin=0,
    ymajorgrids,
    grid style={draw=mygrid, line width=0.5pt},
    axis line style={draw=gray!65},
    tick style={draw=gray!65},
    tick label style={
        font=\scriptsize,
        color=mytext
    },
    tick align=outside,
    tick pos=left,
    label style={font=\small, color=mytext},
    title style={
        font=\small\bfseries,
        color=mytext,
        yshift=-4pt,
    },
    every axis/.append style={
        line width=0.8pt,
    },
]

\nextgroupplot[
    title={(a) Out-of-sight anchors},
    xlabel={Query time (s)},
    ylabel={Number of query anchors},
    xlabel style={yshift=3pt},
    ylabel style={yshift=-4pt},
    xmin=0,
    xmax=600,
    xtick={0,100,200,300,400,500,600},
    ymin=0,
    ymax=85,
    ytick={0,10,20,30,40,50,60,70,80},
]

\addplot[
    ybar,
    bar width=5pt,
    draw=white,
    fill=myteal,
]
table[
    x=time,
    y=count,
    col sep=space,
]
{data/evaluation_set_statistics/query_time.dat};

\addplot[
    domain=0:120,
    samples=2,
    dashed,
    draw=mymuted,
    forget plot,
] ({150}, {x});

\addplot[
    domain=0:120,
    samples=2,
    dashed,
    draw=mymuted,
    forget plot,
] ({300}, {x});

\node[
    anchor=south west,
    fill=myteallight,
    inner sep=2pt,
    font=\scriptsize\bfseries,
    text=mytext,
] at (rel axis cs:0.01,1.08)
{EARLY $\cdot$ MID $\cdot$ LATE};

\node[font=\scriptsize\bfseries, text=mytext]
    at (axis cs:75,82) {Early};
\node[font=\scriptsize\bfseries, text=mytext]
    at (axis cs:75,78) {(333)};

\node[font=\scriptsize\bfseries, text=mytext]
    at (axis cs:225,82) {Mid};
\node[font=\scriptsize\bfseries, text=mytext]
    at (axis cs:225,78) {(333)};

\node[font=\scriptsize\bfseries, text=mytext]
    at (axis cs:450,82) {Late};
\node[font=\scriptsize\bfseries, text=mytext]
    at (axis cs:450,78) {(334)};

\nextgroupplot[
    title={(b) Out-of-sight horizon},
    xlabel={Out-of-sight horizon (s)},
    ylabel={Number of query anchors},
    xlabel style={yshift=3pt},
    ylabel style={yshift=-4pt},
    xmin=0,
    xmax=90,
    xtick={0,10,20,30,40,50,60,70,80,90},
    ymin=0,
    ymax=140,
    ytick={0,20,40,60,80,100,120},
]

\addplot[
    ybar,
    bar width=5pt,
    draw=white,
    fill=myteal,
]
table[
    x=horizon,
    y=count,
    col sep=space,
]
{data/evaluation_set_statistics/horizon.dat};

\addplot[
    domain=0:140,
    samples=2,
    dashed,
    draw=mymuted,
    forget plot,
] ({10.5}, {x});

\addplot[
    domain=0:140,
    samples=2,
    dashed,
    draw=mymuted,
    forget plot,
] ({30.5}, {x});

\node[
    anchor=south west,
    fill=myteallight,
    inner sep=2pt,
    font=\scriptsize\bfseries,
    text=mytext,
] at (rel axis cs:0.01,1.08)
{SHORT $\cdot$ MEDIUM $\cdot$ LONG MEMORY};

\node[font=\scriptsize\bfseries, text=mytext]
    at (axis cs:5.5,135) {Short};
\node[font=\scriptsize\bfseries, text=mytext]
    at (axis cs:5.5,128) {(333)};

\node[font=\scriptsize\bfseries, text=mytext]
    at (axis cs:20.5,135) {Medium};
\node[font=\scriptsize\bfseries, text=mytext]
    at (axis cs:20.5,128) {(333)};

\node[font=\scriptsize\bfseries, text=mytext]
    at (axis cs:60,135) {Long};
\node[font=\scriptsize\bfseries, text=mytext]
    at (axis cs:60,128) {(334)};

\nextgroupplot[
    title={(c) Target moves},
    xlabel={Number of target moves before query},
    ylabel={Number of query anchors},
    xlabel style={yshift=3pt},
    ylabel style={yshift=-4pt},
    xmin=0.4,
    xmax=11.6,
    xtick={1,2,3,4,5,6,7,8,9,10,11},
    ymax=430,
]

\addplot[
    ybar,
    bar width=8pt,
    draw=white,
    fill=myteal,
    nodes near coords,
    every node near coord/.append style={
        font=\scriptsize,
        text=mytext,
        yshift=1pt,
    },
]
table[
    x=moves,
    y=count,
    col sep=space,
]
{data/evaluation_set_statistics/movement.dat};

\node[
    anchor=south west,
    fill=myteallight,
    inner sep=2pt,
    font=\scriptsize\bfseries,
    text=mytext,
] at (rel axis cs:0.01,1.08)
{TARGET MOVES BEFORE QUERY};

\nextgroupplot[
    title={(d) Visible anchors},
    xlabel={Query time (s)},
    ylabel={Number of query anchors},
    xlabel style={yshift=3pt},
    ylabel style={yshift=-4pt},
    xmin=0,
    xmax=600,
    xtick={0,100,200,300,400,500,600},
    ymin=0,
    ymax=85,
    ytick={0,10,20,30,40,50,60,70,80},
]

\addplot[
    ybar,
    bar width=5pt,
    draw=white,
    fill=myteal,
]
table[
    x=time,
    y=count,
    col sep=space,
]
{data/evaluation_set_statistics/visible_query_time.dat};

\addplot[
    domain=0:120,
    samples=2,
    dashed,
    draw=mymuted,
    forget plot,
] ({150}, {x});

\addplot[
    domain=0:120,
    samples=2,
    dashed,
    draw=mymuted,
    forget plot,
] ({300}, {x});

\node[
    anchor=south west,
    fill=myteallight,
    inner sep=2pt,
    font=\scriptsize\bfseries,
    text=mytext,
] at (rel axis cs:0.01,1.08)
{EARLY $\cdot$ MID $\cdot$ LATE};

\node[font=\scriptsize\bfseries, text=mytext]
    at (axis cs:75,82) {Early};
\node[font=\scriptsize\bfseries, text=mytext]
    at (axis cs:75,78) {(333)};

\node[font=\scriptsize\bfseries, text=mytext]
    at (axis cs:225,82) {Mid};
\node[font=\scriptsize\bfseries, text=mytext]
    at (axis cs:225,78) {(333)};

\node[font=\scriptsize\bfseries, text=mytext]
    at (axis cs:450,82) {Late};
\node[font=\scriptsize\bfseries, text=mytext]
    at (axis cs:450,78) {(334)};

\end{groupplot}

\end{tikzpicture}
\end{adjustbox}

\caption{\textbf{Evaluation-set statistics.}
Distribution of the $1{,}000$ out-of-sight query anchors over query time (\textbf{a}), out-of-sight horizon (\textbf{b}), and number of times the target was moved before the query (\textbf{c}), together with the query-time distribution of the $1{,}000$ visible control anchors (\textbf{d}). Dashed lines mark the temporal and horizon stratification boundaries used for sampling.}
\label{fig:eval_set_stat}

\phantomsubcaption\label{fig:oos_query_time}
\phantomsubcaption\label{fig:oos_horizon}
\phantomsubcaption\label{fig:target_moves}
\phantomsubcaption\label{fig:visible_query_time}

\end{figure*}
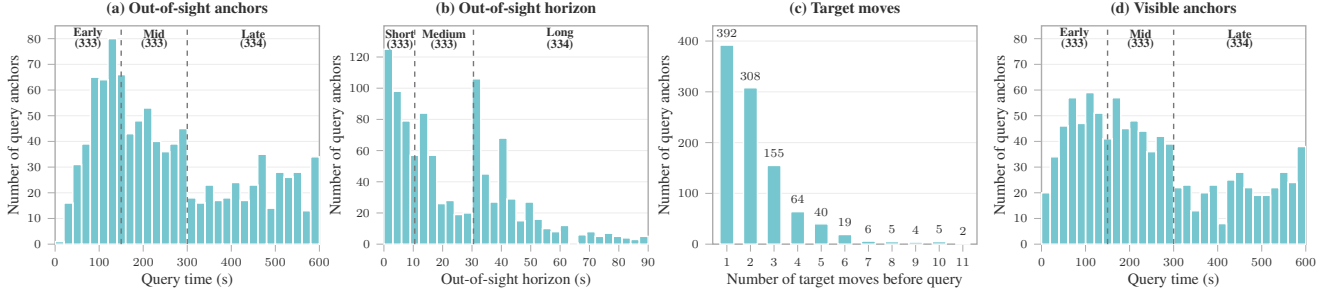

\par\noindent
\textbf{Validation.}
We visually inspected the inferred visibility tracks using an independent human annotation pass. Across $4{,}102$ scored object marks from $344$ frames spanning $30$ videos, the tracks achieved $83.5\%$ accuracy, with most errors arising from the \texttt{visually\_unconfirmed} state. Full validation details are provided in Supp.~\ref{sec:Supplementary_validation}.

\subsection{Q\&A Generation and Statistics}
\label{sec:vqa}
\par\noindent
\textbf{Candidate query anchor selection.}
As summarized in the bottom panel of \cref{fig:Benchmark construction pipeline}, we first select valid query anchors and then instantiate them into Q\&A pairs.
From the constructed visibility tracks, we enumerate valid query anchors $(o,T_q)$ only when the target is explicitly labeled \texttt{out\_\allowbreak of\_\allowbreak view} or \texttt{occluded} at $T_q$. We exclude \texttt{visually\_\allowbreak unconfirmed} states, which often arise from transient dynamic occlusions by hands or other objects, to favor stable out-of-sight periods that more directly test retention of the target's latent spatial state. We retain anchors whose target has a unique, human-readable name.

\par\noindent
\textbf{Q\&A instantiation.}
For a retained query anchor $(o,T_q)$, we instantiate all question types using fixed natural-language templates with placeholders. For example,
``At \texttt{[TIME]}, is the previously moved \texttt{[OBJECT]} visible in the current frame?''
where \texttt{[OBJECT]} denotes $o$ and \texttt{[TIME]} denotes $T_q$. The complete templates are provided in
Supp.~\ref{sec:QA_template}.

Each question is formatted as multiple choice, we then derive the ground-truth answer from visibility tracks, object annotation, and camera poses, and construct distractors using task-specific rules. Temporal distractors are sampled from bins at increasing temporal distances from the ground truth: near ($\pm$1--2\,s), medium ($\pm$3--4\,s), far ($\pm$5--6\,s), and very far ($\pm$7--30\,s), prioritizing timestamps associated with other visibility or movement events. For scene-localization questions, distractors are plausible alternative locations. Since 63.2\% of placements occur on counters, using a single \textit{counter} category would make many questions too coarse and heavily skew the answer distribution. We therefore use fixture categories directly in general, but when the target was last placed on a counter, we instead distinguish counter areas using nearby landmarks (e.g. counter area next to the microwave). For 3D spatial questions, the predefined direction and distance categories directly define the answer options. Full construction details are in Supp.~\ref{sec:Supplementary_vqa_construction}.

\par\noindent
\textbf{Evaluation-set selection and statistics.}
From the candidate pool, we select $1{,}000$ out-of-sight query anchors spanning diverse participants, videos, objects, out-of-sight durations and causes, and answer classes. We prioritize anchors with clear object relocations and reliable visibility evidence. 

We approximately balance the anchors across a $3\times3$ stratification defined by query time $T_q$ (\emph{early}: $0$--$149$\,s, \emph{middle}: $150$--$299$\,s, \emph{late}: $300$--$600$\,s) and out-of-sight horizon $h(o,T_q)$ (\emph{short}: $2$--$10$\,s, \emph{medium}: $11$--$30$\,s, \emph{long}: $>30$\,s). Figs.~\ref{fig:oos_query_time} and~\ref{fig:oos_horizon} show the balanced marginal distributions over query time and out-of-sight horizon, respectively. Fig.~\ref{fig:target_moves} further characterizes spatial-update complexity by the number of target relocations before $T_q$: 70\% of anchors involve one or two target moves, while the long tail extends to 11 moves. 

The resulting set spans 135 videos, nine participants, nine kitchens, 581 target instances, and 561 reference-object instances. At query time, 900 targets are outside the camera's field of view and 100 are geometrically occluded. Expanding each anchor into eight question types produces $8{,}000$ out-of-sight questions, with balanced answer classes for the four 3D spatial question types.

Because these anchors all have negative visibility labels, we additionally sample $1{,}000$ visible anchors as positive controls for the visibility question. These anchors contain previously moved objects that are visible at $T_q$ and are also evenly distributed across the early, middle, and late query-time groups. Their query-time distribution is shown in Fig.~\ref{fig:visible_query_time}. The final evaluation set contains $9{,}000$ questions.

\par\noindent
\textbf{Quality control.}
We manually inspect all $1{,}000$ sampled out-of-sight anchors, verifying from the relevant video evidence that the target is out of sight at query time and that the ground truth can be determined reliably. Ambiguous samples are discarded and replaced from the candidate pool.

\begin{table*}[t]
    \centering
    \caption{
      \textbf{VLM accuracy (\%) on \textsc{Beyond3D}.}
      Performance of VLMs across question types under \textit{Text only}, \textit{Video + Text} and \textit{Last Frame Only + Text} input settings.
      \textbf{Bold} entries indicate the best-performing model for the respective question type.
    }
    \vspace{-10pt}
    \label{tab:main_results}

    \setlength{\tabcolsep}{1.5pt}
    \renewcommand{\arraystretch}{1.10}

    \fontsize{14}{17}\selectfont
    \resizebox{\textwidth}{!}{
    \begin{tabular}{@{}lcccccccccc@{}}
        \toprule
        \textbf{Model}
        & \textbf{Size}
        & \textbf{Macro Avg.}
        & \multicolumn{1}{c}{\textbf{Visual Grounding}}
        & \multicolumn{2}{c}{\textbf{Temporal Grounding}}
        & \multicolumn{1}{c}{\textbf{Scene Localization}}
        & \multicolumn{4}{c}{\textbf{3D Spatial Perception}}
        \\

        \cmidrule(lr){4-4}
        \cmidrule(lr){5-6}
        \cmidrule(lr){7-7}
        \cmidrule(lr){8-11}

        &
        &
        &
        \makecell[c]{\textbf{Visibility}\\\textbf{Check}}
        &
        \makecell[c]{\textbf{Last Visible}\\\textbf{Time}}
        &
        \makecell[c]{\textbf{Last Placement}\\\textbf{Time}}
        &
        \makecell[c]{\textbf{Nearest}\\\textbf{Fixture}}
        &
        \makecell[c]{\textbf{Object--Camera}\\\textbf{Direction}}
        &
        \makecell[c]{\textbf{Object--Camera}\\\textbf{Distance}}
        &
        \makecell[c]{\textbf{Object--Object}\\\textbf{Direction}}
        &
        \makecell[c]{\textbf{Object--Object}\\\textbf{Distance}}
        \\

        \textit{No. of questions}
        &
        &
        9000
        & 2000
        & 1000
        & 1000
        & 1000
        & 1000
        & 1000
        & 1000
        & 1000
        \\
        \midrule

        \rowcolor{black!10}
        \textit{Random guessing}
        & --
        & 29.7
        & 50.0
        & 20.0
        & 20.0
        & 22.7
        & 25.0
        & 33.3
        & 33.3
        & 33.3
        \\

        \midrule
        \rowcolor{blue!8}
        \multicolumn{11}{c}{\textbf{Text Only}}
        \\
        \midrule

        \rowcolor{blue!4}
        \multicolumn{11}{@{}l@{}}{\textit{General-Purpose Models}}
        \\

        Qwen-3.6~\cite{qwen2026qwen36}
        & 35B-A3B
        & 31.0 & 50.0 & 20.9 & 21.2 & 33.4 & 24.6 & 32.5 & 32.6 & 33.0
        \\

        Qwen-3.6~\cite{qwen2026qwen36}
        & 27B
        & 31.7 & 49.5 & 17.8 & 19.0 & 40.2 & 25.0 & 33.6 & 34.3 & 34.1
        \\

        Qwen-3.5~\cite{qwen2026qwen35}
        & 9B
        & 31.5 & 50.0 & 21.1 & 21.8 & 36.5 & 25.9 & 32.6 & 32.0 & 32.1
        \\

        Qwen-3-VL~\cite{bai2025qwen3vl}
        & 8B
        & 30.2 & 50.4 & 19.5 & 20.4 & 25.1 & 25.6 & 33.8 & 34.0 & 32.7
        \\

        InternVL-3.5~\cite{wang2025internvl35}
        & 8B
        & 30.8 & 50.2 & 21.5 & 20.3 & 27.2 & 25.2 & 33.2 & 35.9 & 32.8
        \\

        \addlinespace

        \rowcolor{blue!4}
        \multicolumn{11}{@{}l@{}}{\textit{Specialized 3D Models}}
        \\

        VLM-3R~\cite{fan2025vlm3r}
        & 7B
        & 31.0 & 50.0 & 21.3 & 19.4 & 34.5 & 24.6 & 31.8 & 33.7 & 33.0
        \\

        Spatial-MLLM~\cite{wu2025spatialmllm}
        & 6B
        & 30.0 & 49.4 & 22.4 & 19.8 & 22.2 & 25.1 & 32.7 & 34.1 & 34.0
        \\

        Cambrian-P~\cite{yang2026cambrianp}
        & 7B
        & 31.9 & 51.1 & 19.7 & 18.0 & 41.3 & 25.0 & 33.4 & 33.5 & 33.3
        \\

        SenseNova-SI~\cite{cai2025sensenovasi}
        & 8B
        & 29.0 & 51.2 & 18.0 & 18.1 & 20.7 & 26.0 & 33.7 & 31.3 & 33.1
        \\

        \midrule
        \rowcolor{blue!8}

        \multicolumn{11}{c}{\textbf{Last Frame Only + Text}}
        \\
        \midrule

        \rowcolor{blue!4}
        \multicolumn{11}{@{}l@{}}{\textit{General-Purpose Models}}
        \\

        Qwen-3.6~\cite{qwen2026qwen36}
        & 35B-A3B
        & 33.0 & 72.9 & 14.5 & 16.6 & 27.1 & 27.0 & 35.3 & 35.2 & 35.0
        \\

        Qwen-3.6~\cite{qwen2026qwen36}
        & 27B
        & 35.2 & \textbf{76.5} & 17.8 & 18.3 & 29.0 & 29.7 & 38.0 & 36.5 & 35.9
        \\

        Qwen-3.5~\cite{qwen2026qwen35}
        & 9B
        & 32.4 & 64.3 & 17.9 & 21.2 & 25.3 & 26.6 & 33.6 & 34.8 & 35.1
        \\

        Qwen-3-VL~\cite{bai2025qwen3vl}
        & 8B
        & 31.8 & 73.4 & 14.5 & 18.0 & 21.4 & 26.8 & 35.1 & 32.8 & 32.7
        \\

        InternVL-3.5~\cite{wang2025internvl35}
        & 8B
        & 32.1 & 71.3 & 14.9 & 16.5 & 24.5 & 27.9 & 34.8 & 33.3 & 34.0
        \\

        \addlinespace

        \rowcolor{blue!4}
        \multicolumn{11}{@{}l@{}}{\textit{Specialized 3D Models}}
        \\

        VLM-3R~\cite{fan2025vlm3r}
        & 7B
        & 33.6 & 68.2 & 21.5 & 21.8 & 27.7 & 28.1 & 33.6 & 33.8 & 34.1
        \\

        Spatial-MLLM~\cite{wu2025spatialmllm}
        & 6B
        & 31.9 & 60.1 & 22.3 & 20.0 & 23.8 & 25.5 & 35.6 & 33.5 & 33.9
        \\

        Cambrian-P~\cite{yang2026cambrianp}
        & 7B
        & 32.9 & 69.5 & 18.7 & 18.2 & 31.0 & 25.4 & 33.4 & 33.9 & 33.3
        \\

        SenseNova-SI~\cite{cai2025sensenovasi}
        & 8B
        & 32.7 & 74.2 & 17.1 & 20.4 & 19.8 & 25.8 & 37.7 & 31.4 & 35.3
        \\

        \midrule
        \rowcolor{blue!8}
                \multicolumn{11}{c}{\textbf{Video + Text}}
        \\
        \midrule

        \rowcolor{blue!4}
        \multicolumn{11}{@{}l@{}}{\textit{General-Purpose Models}}
        \\

        Qwen-3.6~\cite{qwen2026qwen36}
        & 35B-A3B
        & 39.6 & 66.3 & 23.3 & 29.4 & \textbf{51.3} & 33.7 & 35.9 & 41.1 & 35.6
        \\

        Qwen-3.6~\cite{qwen2026qwen36}
        & 27B
        & \textbf{42.2} & 68.4 & \textbf{30.4} & \textbf{39.4} & 50.0 & 34.6 & 34.9 & \textbf{43.6} & 36.2
        \\

        Qwen-3.5~\cite{qwen2026qwen35}
        & 9B
        & 38.6 & 59.3 & 28.2 & 30.5 & 47.9 & 33.4 & 39.2 & 34.7 & 35.8
        \\

        Qwen-3-VL~\cite{bai2025qwen3vl}
        & 8B
        & 36.0 & 59.8 & 21.3 & 25.7 & 48.2 & 32.7 & 33.6 & 31.5 & 35.2
        \\

        InternVL-3.5~\cite{wang2025internvl35}
        & 8B
        & 35.8 & 60.4 & 21.8 & 20.7 & 38.9 & 31.4 & \textbf{42.6} & 35.8 & 34.9
        \\

        \addlinespace

        \rowcolor{blue!4}
        \multicolumn{11}{@{}l@{}}{\textit{Specialized 3D Models}}
        \\

        VLM-3R~\cite{fan2025vlm3r}
        & 7B
        & 37.2 & 58.8 & 23.9 & 23.2 & 49.8 & \textbf{35.6} & 34.4 & 35.6 & 36.5
        \\

        Spatial-MLLM~\cite{wu2025spatialmllm}
        & 6B
        & 31.6 & 53.3 & 22.8 & 19.2 & 29.5 & 24.6 & 34.8 & 34.0 & 34.8
        \\

        Cambrian-P~\cite{yang2026cambrianp}
        & 7B
        & 33.5 & 52.2 & 20.0 & 17.5 & 50.5 & 27.3 & 33.4 & 33.9 & 33.2
        \\

        SenseNova-SI~\cite{cai2025sensenovasi}
        & 8B
        & 35.4 & 57.9 & 20.6 & 22.9 & 44.4 & 29.8 & 32.9 & 34.9 & \textbf{39.8}
        \\

        \bottomrule
    \end{tabular}%
    }
\end{table*}

\section{Experiments}
\label{sec:exp}
\subsection{Experimental Setup}
\par\noindent
\textbf{Evaluated models.}
We evaluate nine recent VLMs spanning general-purpose and spatially specialized models.
The general-purpose models are Qwen-3.6~\cite{qwen2026qwen36} in its 35B-A3B and 27B variants, Qwen-3.5~\cite{qwen2026qwen35}, Qwen-3-VL~\cite{bai2025qwen3vl}, and InternVL-3.5~\cite{wang2025internvl35}. The spatially specialized models are VLM-3R~\cite{fan2025vlm3r}, Spatial-MLLM~\cite{wu2025spatialmllm}, Cambrian-P~\cite{yang2026cambrianp}, and SenseNova-SI~\cite{cai2025sensenovasi}. We use the authors' publicly released checkpoints and official inference implementations without task-specific fine-tuning. All model inference was conducted on an HPC cluster, using a single NVIDIA A100 or NVIDIA RTX PRO 6000 GPU per run, with 80 GB and 96 GB of GPU memory, respectively.
\par\noindent
\textbf{Evaluation protocol.}
We report per-question multiple-choice accuracy and macro-average accuracy, giving each question type equal weight despite \textit{Visibility Check} having twice as many questions compared to others. Each model receives the 1\,fps video prefix up to the query time $T_q$. For models with shorter context limits, frames are uniformly subsampled to fit the available context. Visual preprocessing and the full prompts are provided in Supp.~\ref{sec:Visual_input_preprocessing} and~\ref{sec:Inference_prompt}.

\subsection{Main Results}
We present the performance of different models in Table~\ref{tab:main_results} evaluated in three settings: (i) \textit{Text only}, where the model receives only the question and options; (ii) \textit{Last Frame Only + Text}, where the frame at the query time is additionally provided; and (iii) \textit{Video + Text}, where the video prefix is provided instead of the last frame.

\par\noindent
\textbf{Text-only performance is near chance.}
Without video, models perform close to random guessing on nearly all question types. The main exception is \textit{Nearest Fixture}, where several models achieve higher accuracy. This reflects object--fixture priors acquired during pretraining, such as associating a milk carton with a fridge.

\par\noindent
\textbf{Video evidence improves on text-only performance.}
Video improves macro accuracy for every model by 1.6--10.5\%, with the largest average gains on \textit{Visibility Check} (+9.4\%) and \textit{Nearest Fixture} (+14.4\%). Gains are less consistent for temporal grounding and smaller for spatial tasks. Despite these gains, overall performance remains low.

\par\noindent
\textbf{Difficulty varies across tasks.}
Relative to chance, VLMs perform best on \textit{Nearest Fixture} (45.6\% vs.\ 22.7\%), followed by visual grounding (59.6\% vs.\ 50.0\%). In contrast, temporal grounding and spatial reasoning are substantially harder, exceeding chance by only 4.5\% and 3.6\% on average, respectively. Thus, models are better at recovering coarse semantic location than at precisely grounding past events or reasoning about remembered positions in 3D.

\par\noindent
\textbf{General-purpose models vs.\ spatially specialized models.}
General-purpose models perform better overall, led by Qwen-3.6-27B at 42.2\% macro accuracy. This is not a scale effect: Qwen-3.5-9B (38.6\%) outperforms all comparably sized spatially specialized models (31.6--37.2\%). On the 3D spatial tasks, these models show no consistent advantage and remain modestly above random. We hypothesize that this reflects a mismatch between their spatial specialization and our setting: existing 3D training mainly targets scene geometry and viewpoint-induced spatial changes, while \textsc{Beyond3D} additionally requires upstream temporal grounding of object relocations and retaining the spatial state while the object is not viewable. Their weaker temporal grounding and evidence retrieval ability creates a bottleneck that limits the benefit of 3D reasoning. More restrictive context budgets for several spatially specialized models may contribute to this gap.

\par\noindent
\textbf{Current-frame evidence is insufficient for out-of-sight reasoning.} Providing only the query frame substantially improves \textit{Visibility Check}, reaching 70.0\% on average, but leaves temporal grounding, scene localization, and 3D spatial reasoning near chance. In contrast, full-video input substantially improves tasks that require recovering the target's earlier state: \textit{Last Visible Time} increases from 17.7\% to 23.6\%, \textit{Last Placement Time} from 19.0\% to 25.4\%, and \textit{Nearest Fixture} from 25.5\% to 45.6\%. This shows that the benchmark separates instantaneous visibility perception from reasoning over latent object states that must be recovered from video history.

\subsection{Diagnosing Failure Modes}
Unless otherwise stated, all subsequent analyses use the \textit{Video + Text} setting, corresponding to the full-video benchmark setting.
\input{figures/temporal_performance_curves}
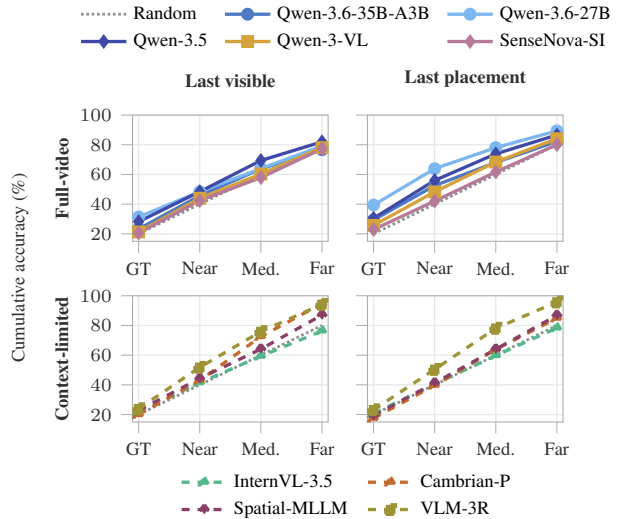
\begin{figure}[t]
\centering

\definecolor{fvA}{HTML}{4E79C5}
\definecolor{fvB}{HTML}{7DB7F0}
\definecolor{fvC}{HTML}{3F4AA8}
\definecolor{fvD}{HTML}{D5A23B}
\definecolor{fvE}{HTML}{B57A9A}

\definecolor{clA}{HTML}{53B88B}
\definecolor{clB}{HTML}{C56A2D}
\definecolor{clC}{HTML}{8A3F69}
\definecolor{clD}{HTML}{9D9436}

\definecolor{tempgrid}{HTML}{E8E8E8}
\definecolor{temptext}{HTML}{303030}
\definecolor{tempbaseline}{HTML}{8C8C8C}

\begin{adjustbox}{width=\columnwidth}
\begin{tikzpicture}

\pgfplotsset{
    compact temporal/.style={
        width=0.5\columnwidth,
        height=3.25cm,
        xmin=-0.1,
        xmax=3.1,
        ymin=15,
        ymax=100,
        xtick={0,1,2,3},
        xticklabels={GT,Near,Med.,Far},
        ytick={20,40,60,80,100},
        xmajorgrids,
        ymajorgrids,
        grid style={draw=tempgrid,line width=0.5pt},
        axis line style={draw=gray!65},
        tick style={draw=gray!65},
        tick pos=left,
        tick align=outside,
        tick label style={
            font=\scriptsize,
            color=temptext
        },
        title style={
            font=\scriptsize\bfseries,
            color=temptext,
            yshift=1pt
        }
    }
}

\begin{axis}[
    compact temporal,
    name=p1,
    title={Last visible},
    legend to name=fulllegend,
    legend columns=3,
    legend style={
        draw=none,
        font=\scriptsize,
        cells={anchor=west},
        /tikz/every even column/.append style={column sep=4pt}
    }
]

\addplot[
    color=tempbaseline,
    densely dotted,
    line width=1pt
] coordinates {(0,20) (1,40) (2,60) (3,80)};
\addlegendentry{Random}

\addplot[color=fvA,line width=1.25pt,mark=*,mark size=1.8pt]
table[x=x,y=qwen36_35b_a3b,col sep=tab]
{data/cumulative_accuracy_over_temporal_distance/full_video_last_visible.dat};
\addlegendentry{Qwen-3.6-35B-A3B}

\addplot[color=fvB,line width=1.25pt,mark=*,mark size=1.8pt]
table[x=x,y=qwen36_27b,col sep=tab]
{data/cumulative_accuracy_over_temporal_distance/full_video_last_visible.dat};
\addlegendentry{Qwen-3.6-27B}

\addplot[color=fvC,line width=1.25pt,mark=diamond*,mark size=1.8pt]
table[x=x,y=qwen35_9b,col sep=tab]
{data/cumulative_accuracy_over_temporal_distance/full_video_last_visible.dat};
\addlegendentry{Qwen-3.5}

\addplot[color=fvD,line width=1.25pt,mark=square*,mark size=1.8pt]
table[x=x,y=qwen3_vl_8b,col sep=tab]
{data/cumulative_accuracy_over_temporal_distance/full_video_last_visible.dat};
\addlegendentry{Qwen-3-VL}

\addplot[color=fvE,line width=1.25pt,mark=diamond*,mark size=1.8pt]
table[x=x,y=sensenova_si,col sep=tab]
{data/cumulative_accuracy_over_temporal_distance/full_video_last_visible.dat};
\addlegendentry{SenseNova-SI}

\end{axis}

\begin{axis}[
    compact temporal,
    name=p2,
    at={(p1.south east)},
    anchor=south west,
    xshift=0.52cm,
    yticklabels={,,,,},
    title={Last placement}
]

\addplot[
    color=tempbaseline,
    densely dotted,
    line width=1pt,
    forget plot
] coordinates {(0,20) (1,40) (2,60) (3,80)};

\addplot[color=fvA,line width=1.25pt,mark=*,mark size=1.8pt]
table[x=x,y=qwen36_35b_a3b,col sep=tab]
{data/cumulative_accuracy_over_temporal_distance/full_video_last_placement.dat};

\addplot[color=fvB,line width=1.25pt,mark=*,mark size=1.8pt]
table[x=x,y=qwen36_27b,col sep=tab]
{data/cumulative_accuracy_over_temporal_distance/full_video_last_placement.dat};

\addplot[color=fvC,line width=1.25pt,mark=diamond*,mark size=1.8pt]
table[x=x,y=qwen35_9b,col sep=tab]
{data/cumulative_accuracy_over_temporal_distance/full_video_last_placement.dat};

\addplot[color=fvD,line width=1.25pt,mark=square*,mark size=1.8pt]
table[x=x,y=qwen3_vl_8b,col sep=tab]
{data/cumulative_accuracy_over_temporal_distance/full_video_last_placement.dat};

\addplot[color=fvE,line width=1.25pt,mark=diamond*,mark size=1.8pt]
table[x=x,y=sensenova_si,col sep=tab]
{data/cumulative_accuracy_over_temporal_distance/full_video_last_placement.dat};

\end{axis}

\begin{axis}[
    compact temporal,
    name=p3,
    at={(p1.south west)},
    anchor=north west,
    yshift=-0.72cm,
    legend to name=contextlegend,
    legend columns=2,
    legend style={
        draw=none,
        font=\scriptsize,
        cells={anchor=west},
        /tikz/every even column/.append style={column sep=5pt}
    }
]

\addplot[
    color=tempbaseline,
    densely dotted,
    line width=1pt,
    forget plot
] coordinates {(0,20) (1,40) (2,60) (3,80)};

\addplot[color=clA,line width=1.25pt,mark=triangle*,mark size=1.9pt,dashed]
table[x=x,y=internvl35_8b,col sep=tab]
{data/cumulative_accuracy_over_temporal_distance/context_limited_last_visible.dat};
\addlegendentry{InternVL-3.5}

\addplot[color=clB,line width=1.25pt,mark=triangle*,mark size=1.9pt,dashed]
table[x=x,y=cambrian_p_7b,col sep=tab]
{data/cumulative_accuracy_over_temporal_distance/context_limited_last_visible.dat};
\addlegendentry{Cambrian-P}

\addplot[color=clC,line width=1.25pt,mark=diamond*,mark size=1.8pt,dashed]
table[x=x,y=spatial_mllm,col sep=tab]
{data/cumulative_accuracy_over_temporal_distance/context_limited_last_visible.dat};
\addlegendentry{Spatial-MLLM}

\addplot[color=clD,line width=1.25pt,mark=square*,mark size=1.8pt,dashed]
table[x=x,y=vlm3r,col sep=tab]
{data/cumulative_accuracy_over_temporal_distance/context_limited_last_visible.dat};
\addlegendentry{VLM-3R}

\end{axis}

\begin{axis}[
    compact temporal,
    name=p4,
    at={(p2.south west)},
    anchor=north west,
    yshift=-0.72cm,
    yticklabels={,,,,}
]

\addplot[
    color=tempbaseline,
    densely dotted,
    line width=1pt,
    forget plot
] coordinates {(0,20) (1,40) (2,60) (3,80)};

\addplot[color=clA,line width=1.25pt,mark=triangle*,mark size=1.9pt,dashed]
table[x=x,y=internvl35_8b,col sep=tab]
{data/cumulative_accuracy_over_temporal_distance/context_limited_last_placement.dat};

\addplot[color=clB,line width=1.25pt,mark=triangle*,mark size=1.9pt,dashed]
table[x=x,y=cambrian_p_7b,col sep=tab]
{data/cumulative_accuracy_over_temporal_distance/context_limited_last_placement.dat};

\addplot[color=clC,line width=1.25pt,mark=diamond*,mark size=1.8pt,dashed]
table[x=x,y=spatial_mllm,col sep=tab]
{data/cumulative_accuracy_over_temporal_distance/context_limited_last_placement.dat};

\addplot[color=clD,line width=1.25pt,mark=square*,mark size=1.8pt,dashed]
table[x=x,y=vlm3r,col sep=tab]
{data/cumulative_accuracy_over_temporal_distance/context_limited_last_placement.dat};

\end{axis}

\path (p1.north east) --
    coordinate[midway] (topcenter)
    (p2.north west);

\path (p3.south east) --
    coordinate[midway] (bottomcenter)
    (p4.south west);

\node[anchor=south,yshift=17pt]
at (topcenter)
{\pgfplotslegendfromname{fulllegend}};

\node[anchor=north,yshift=-12pt]
at (bottomcenter)
{\pgfplotslegendfromname{contextlegend}};

\node[
    rotate=90,
    anchor=south,
    font=\scriptsize,
    text=temptext
] at ([xshift=-1.25cm,yshift=-1.185cm]p1.west)
{Cumulative accuracy (\%)};

\node[
    rotate=90,
    anchor=south,
    font=\scriptsize\bfseries,
    text=temptext
] at ([xshift=-0.72cm]p1.west)
{Full-video};

\node[
    rotate=90,
    anchor=south,
    font=\scriptsize\bfseries,
    text=temptext
] at ([xshift=-0.72cm]p3.west)
{Context-limited};

\end{tikzpicture}
\end{adjustbox}
\vspace{-15pt}
\caption{
\textbf{Cumulative accuracy over temporal distance.}
Accuracy when progressively accepting the rounded ground truth (GT) and timestamp choices up to \emph{Near} ($\pm$1--2\,s), \emph{Medium} ($\pm$3--4\,s), and \emph{Far} ($\pm$5--6\,s).
\textbf{Top:} full-video models; \textbf{Bottom:} context-limited models (InternVL-3.5 / Cambrian-P / VLM-3R: 150 frames, Spatial-MLLM: 64 frames).
\textbf{Left:} last-visible time; \textbf{Right:} last-placement time.
}
\vspace{-5pt}
\label{fig:cumulative_temporal_distance}

\end{figure}
\begin{figure}[t]
\centering

\definecolor{notvis}{HTML}{4C78A8}
\definecolor{vis}{HTML}{7CB7E3}
\definecolor{gridgray}{HTML}{D9D9D9}
\definecolor{spinegray}{HTML}{B5B5B5}
\definecolor{textgray}{HTML}{666666}

\begin{tikzpicture}
\begin{axis}[
    xbar,
    width=0.9\columnwidth,
    height=5.4cm,
    xmin=0,
    xmax=105,
    xtick={0,20,40,60,80,100},
    xlabel={Accuracy (\%)},
    title style={font=\small,yshift=1mm},
    symbolic y coords={
        SenseNova-SI,
        Cambrian-P,
        Spatial-MLLM,
        VLM-3R,
        InternVL-3.5,
        Qwen-3-VL,
        Qwen-3.5,
        Qwen-3.6-27B,
        Qwen-3.6-35B-A3B
    },
    ytick=data,
    yticklabel style={
        font=\scriptsize,
        text width=1.6cm,
        align=right
    },
    xticklabel style={font=\scriptsize},
    label style={font=\scriptsize},
    bar width=4.2pt,
    enlarge y limits=0.08,
    xmajorgrids=true,
    ymajorgrids=false,
    grid style={draw=gridgray,line width=0.35pt},
    axis line style={draw=spinegray,line width=0.5pt},
    tick style={draw=spinegray,line width=0.5pt},
    legend style={
        at={(0.5,1.03)},
        anchor=south,
        legend columns=2,
        draw=none,
        font=\scriptsize,
        /tikz/every even column/.append style={column sep=6pt}
    },
    nodes near coords,
    nodes near coords align={horizontal},
    point meta=x,
    every node near coord/.append style={
        font=\tiny,
        text=black,
        /pgf/number format/fixed,
        /pgf/number format/precision=1
    }
]

\addplot+[
    fill=notvis,
    draw=white,
    line width=0.35pt,
    bar shift=-2.5pt
] coordinates {
    (53.3,SenseNova-SI)
    (12.0,Cambrian-P)
    (18.2,Spatial-MLLM)
    (41.3,VLM-3R)
    (87.8,InternVL-3.5)
    (76.3,Qwen-3-VL)
    (57.1,Qwen-3.5)
    (71.8,Qwen-3.6-27B)
    (48.7,Qwen-3.6-35B-A3B)
};

\addplot+[
    fill=vis,
    draw=white,
    line width=0.35pt,
    bar shift=2.5pt
] coordinates {
    (62.5,SenseNova-SI)
    (92.3,Cambrian-P)
    (83.3,Spatial-MLLM)
    (76.3,VLM-3R)
    (32.9,InternVL-3.5)
    (43.2,Qwen-3-VL)
    (61.4,Qwen-3.5)
    (65.0,Qwen-3.6-27B)
    (83.8,Qwen-3.6-35B-A3B)
};

\legend{Not visible,Visible}

\end{axis}
\end{tikzpicture}
\caption{\textbf{Visibility-state estimation.}
Accuracy in determining whether the queried object is visible at query time,
reported separately for objects that are \emph{not visible} and \emph{visible}.}
\vspace{-15pt}
\label{fig:visibility_state}
\end{figure}
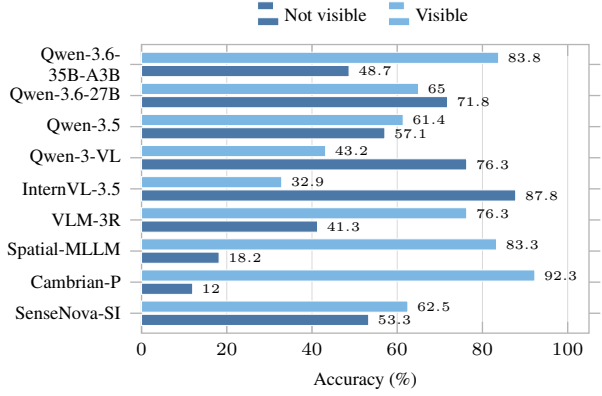
\begin{figure}[t]
\centering

\definecolor{posA}{HTML}{5F86BB}
\definecolor{posB}{HTML}{83A1C9}
\definecolor{posC}{HTML}{A8BDD8}
\definecolor{posD}{HTML}{CAD6E7}

\definecolor{camdistA}{HTML}{91BC8C}
\definecolor{camdistB}{HTML}{ACCCAA}
\definecolor{camdistC}{HTML}{D0E0CE}

\definecolor{objposA}{HTML}{9889BF}
\definecolor{objposB}{HTML}{B2A6CF}
\definecolor{objposC}{HTML}{D0C9E1}

\definecolor{objdistA}{HTML}{E2A354}
\definecolor{objdistB}{HTML}{EBC083}
\definecolor{objdistC}{HTML}{F3D9B5}

\definecolor{predgrid}{HTML}{ECECEC}
\definecolor{predtext}{HTML}{303030}

\begin{adjustbox}{width=\columnwidth}
\begin{tikzpicture}

\begin{groupplot}[
    group style={
        group size=2 by 2,
        horizontal sep=0.6cm,
        vertical sep=1.1cm
    },
    width=0.32\columnwidth,
    height=2.9cm,
    scale only axis,
    xmin=0,
    xmax=100,
    xtick=\empty,
    ymin=-0.5,
    ymax=9.5,
    ytick={0,...,9},
    y dir=reverse,
    xbar stacked,
    /pgf/bar width=4pt,
    xmajorgrids,
    grid style={
        draw=predgrid,
        line width=0.5pt
    },
    axis x line*=bottom,
    axis y line*=left,
    tick style={draw=gray!60},
    tick label style={
        font=\scriptsize,
        color=predtext
    },
    yticklabel style={
        font=\scriptsize
    },
    label style={
        font=\scriptsize,
        color=predtext
    },
    xlabel={}
]

\nextgroupplot[
    yticklabels={
        Ground truth,
        Qwen-3.6-35B-A3B,
        Qwen-3.6-27B,
        Qwen-3.5,
        Qwen-3-VL,
        InternVL-3.5,
        VLM-3R,
        Spatial-MLLM,
        Cambrian-P,
        SenseNova-SI
    }
]

\addplot[fill=posA,draw=white]
table[col sep=tab,x=front_left,y=row]
{data/prediction_distributions/camera_position.dat};

\addplot[fill=posB,draw=white]
table[col sep=tab,x=front_right,y=row]
{data/prediction_distributions/camera_position.dat};

\addplot[fill=posC,draw=white]
table[col sep=tab,x=back_left,y=row]
{data/prediction_distributions/camera_position.dat};

\addplot[fill=posD,draw=white]
table[col sep=tab,x=back_right,y=row]
{data/prediction_distributions/camera_position.dat};

\nextgroupplot[
    ytick=\empty
]

\addplot[fill=camdistA,draw=white]
table[col sep=tab,x=under_1m,y=row]
{data/prediction_distributions/camera_distance.dat};

\addplot[fill=camdistB,draw=white]
table[col sep=tab,x=from_1m_to_1_5m,y=row]
{data/prediction_distributions/camera_distance.dat};

\addplot[fill=camdistC,draw=white]
table[col sep=tab,x=at_least_1_5m,y=row]
{data/prediction_distributions/camera_distance.dat};

\nextgroupplot[
    yticklabels={
        Ground truth,
        Qwen3.6-35B,
        Qwen3.6-27B,
        Qwen3.5,
        Qwen3-VL,
        InternVL3.5,
        VLM-3R,
        Spatial-MLLM,
        Cambrian-P,
        SenseNova-SI
    }
]

\addplot[fill=objposA,draw=white]
table[col sep=tab,x=clock_12_to_4_30,y=row]
{data/prediction_distributions/object_position.dat};

\addplot[fill=objposB,draw=white]
table[col sep=tab,x=clock_4_30_to_7_30,y=row]
{data/prediction_distributions/object_position.dat};

\addplot[fill=objposC,draw=white]
table[col sep=tab,x=clock_7_30_to_12,y=row]
{data/prediction_distributions/object_position.dat};

\nextgroupplot[
    ytick=\empty
]

\addplot[fill=objdistA,draw=white]
table[col sep=tab,x=under_1m,y=row]
{data/prediction_distributions/object_distance.dat};

\addplot[fill=objdistB,draw=white]
table[col sep=tab,x=from_1m_to_1_5m,y=row]
{data/prediction_distributions/object_distance.dat};

\addplot[fill=objdistC,draw=white]
table[col sep=tab,x=at_least_1_5m,y=row]
{data/prediction_distributions/object_distance.dat};

\end{groupplot}

\node[
    anchor=south,
    font=\scriptsize\bfseries
] at ([yshift=12pt]group c1r1.north)
{(a) Camera direction};

\node[
    anchor=south,
    font=\scriptsize\bfseries
] at ([yshift=12pt]group c2r1.north)
{(b) Camera distance};

\node[
    anchor=south,
    font=\scriptsize\bfseries
] at ([yshift=12pt]group c1r2.north)
{(c) Object direction};

\node[
    anchor=south,
    font=\scriptsize\bfseries
] at ([yshift=12pt]group c2r2.north)
{(d) Object distance};

\node[
    anchor=south,
    font=\scriptsize\selectfont
] at ([yshift=2pt]group c1r1.north) {
    \textcolor{posA}{\rule{2.5pt}{2.5pt}}\,FL\,
    \textcolor{posB}{\rule{2.5pt}{2.5pt}}\,FR\,
    \textcolor{posC}{\rule{2.5pt}{2.5pt}}\,BL\,
    \textcolor{posD}{\rule{2.5pt}{2.5pt}}\,BR
};

\node[
    anchor=south,
    font=\scriptsize\selectfont
] at ([yshift=2pt]group c2r1.north) {
    \textcolor{camdistA}{\rule{2.5pt}{2.5pt}}\,$<1$\,
    \textcolor{camdistB}{\rule{2.5pt}{2.5pt}}\,$1$--$1.5$\,
    \textcolor{camdistC}{\rule{2.5pt}{2.5pt}}\,$\geq1.5$ m
};

\node[
    anchor=south,
    font=\scriptsize\selectfont
] at ([yshift=2pt]group c1r2.north) {
    \textcolor{objposA}{\rule{2.5pt}{2.5pt}}\,12--4:30\,
    \textcolor{objposB}{\rule{2.5pt}{2.5pt}}\,4:30--7:30\,
    \textcolor{objposC}{\rule{2.5pt}{2.5pt}}\,7:30--12
};

\node[
    anchor=south,
    font=\scriptsize\selectfont
] at ([yshift=2pt]group c2r2.north) {
    \textcolor{objdistA}{\rule{2.5pt}{2.5pt}}\,$<1$\,
    \textcolor{objdistB}{\rule{2.5pt}{2.5pt}}\,$1$--$1.5$\,
    \textcolor{objdistC}{\rule{2.5pt}{2.5pt}}\,$\geq1.5$ m
};

\end{tikzpicture}
\end{adjustbox}

\caption{\textbf{Prediction distributions for 3D spatial perception.}
Stacked bars show prediction frequencies (\%).
\textbf{(a)} object--camera direction (FL/FR/BL/BR: front/back left/right);
\textbf{(b)} object--camera distance;
\textbf{(c)} camera-aligned object--object direction;
\textbf{(d)} object--object distance.}
\vspace{-15pt}
\label{fig:spatial_prediction}
\end{figure}
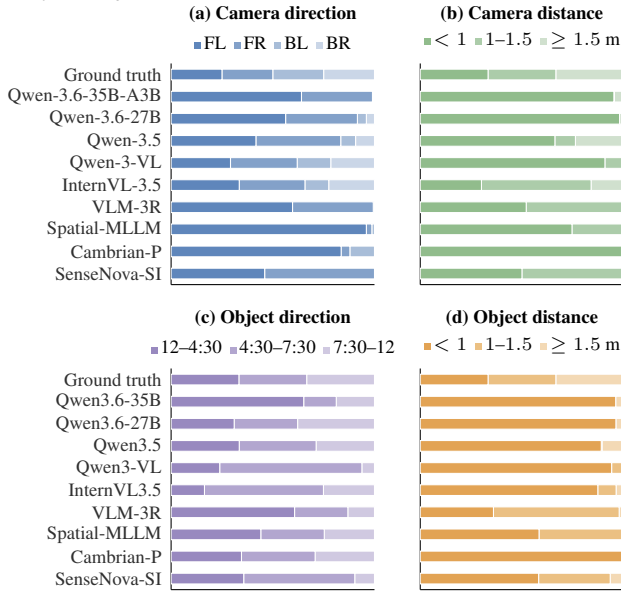
\par\noindent
\textbf{Temporal evidence retrieval is an upstream bottleneck.}
Performance reveals a strong dependence on how long the target remains out of sight (Fig.~\ref{fig:temporal_performance_curves}). Macro average accuracy decreases from 40.4\% for short horizons to 34.8\% for medium and 31.9\% for long horizons. As the horizon grows, the target's last observed state must be retained across more intervening activity, making it increasingly difficult to recover at query time. In contrast, accuracy varies non-monotonically with query time (34.9\%, 38.4\%, and 36.6\% for early, middle, and late queries). Later queries provide more scene evidence but also longer histories, which may explain the middle-query peak. Thus, out-of-sight horizon drives difficulty more than video length.

We next examine one potential upstream source of these errors, locating the interaction that determines the target's latest state (Fig.~\ref{fig:cumulative_temporal_distance}). Since the four distractors lie in bins increasingly distant from the ground truth, the plot progressively counts predictions from farther bins as correct, causing random-choice accuracy to rise from 20\% to 80\%. If models found the relevant event but missed its exact timestamp, their curves would rise faster than this baseline. Most remain close to random, indicating weak preference for the correct temporal region and suggesting that models often retrieve another plausible visibility or movement event.

\par\noindent
\textbf{Out-of-sight state recognition is unreliable.}
Fig.~\ref{fig:visibility_state} evaluates the \textit{Visibility Check} separately for visible and out-of-sight targets. A robust model should perform well in both conditions, yet several models show strong asymmetries. InternVL-3.5 correctly predicts \textit{not visible} for $87.8\%$ of out-of-sight targets but \textit{visible} for only $32.9\%$ of visible targets, while Cambrian-P shows the opposite bias. These opposing errors suggest model-specific visibility biases rather than uniformly weak visual recognition: some models tend to assume that previously observed objects remain absent, whereas others over-rely on current visual evidence. Since downstream questions require reasoning about an object specifically when it is no longer visible, such failures can corrupt the spatial state used for subsequent predictions.

\par\noindent
\textbf{Spatial predictions exhibit strong biases.}
As shown in Fig.~\ref{fig:spatial_prediction}, despite balanced ground-truth answer distributions for each 3D task, camera-direction predictions are skewed toward locations in front of the camera for every model, reaching 98.7\% for Spatial-MLLM~\cite{wu2025spatialmllm}, while most models concentrate both distance tasks on the nearest bin, with Cambrian-P~\cite{yang2026cambrianp} assigning 100\% and 99.4\% of its \textit{Object--Camera Distance} and \textit{Object--Object Distance} predictions there, respectively. One explanation is that, when the remembered target state is uncertain, models fall back on spatial configurations supported by the direct visual evidence, favoring objects that are in front of and close to the camera.


\usepgfplotslibrary{fillbetween,groupplots}
\pgfplotsset{compat=1.18}

\definecolor{TCQCharcoal}{HTML}{303030}
\definecolor{TCQMuted}{HTML}{777777}
\definecolor{TCQLight}{HTML}{D9D9D9}
\definecolor{TCQGrid}{HTML}{ECECEC}
\definecolor{TCQTeal}{HTML}{75C6CE}
\definecolor{TCQTealLight}{HTML}{DFF3F5}
\definecolor{TCQQwenBlue}{HTML}{56B4E9}
\definecolor{TCQQFiveA}{HTML}{477DB5}
\definecolor{TCQQFiveB}{HTML}{82BE8A}
\definecolor{TCQQFiveC}{HTML}{AE9ACC}
\definecolor{TCQQFiveD}{HTML}{F0A556}
\definecolor{TCQNegativeArea}{HTML}{F7F7F7}

\newcommand{\TCQSmall}{%
  \fontsize{6.5}{7.2}\selectfont
}

\newcommand{\TCQQuestionPoint}[5]{%
  \addplot[
    only marks,
    color=#1,
    mark=*,
    mark size=2.2pt,
    mark options={
      draw=#1,
      fill=#1
    },
    error bars/.cd,
      x dir=both,
      x explicit,
      error mark options={
        color=#1,
        mark size=1.4pt,
        line width=0.6pt
      },
      error bar style={
        color=#1,
        line width=0.8pt
      }
  ] table[
    row sep=\\,
    x=delta,
    y=y,
    x error minus=minus,
    x error plus=plus
  ] {
    delta y minus plus\\
    #2 #5 #3 #4\\
  };
}

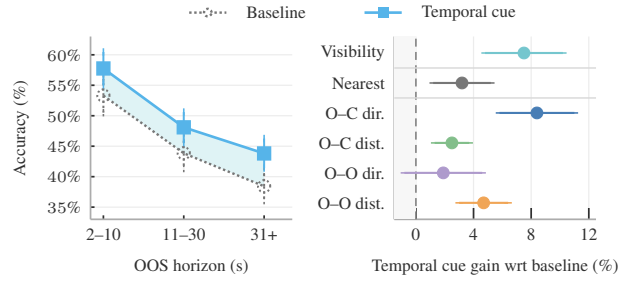
\begin{figure}[t]
\centering

\begin{tikzpicture}

\begin{groupplot}[
  group style={
    group size=2 by 1,
    horizontal sep=1.35cm
  },
  width=4.25cm,
  height=4cm,
  axis line style={
    draw=gray!65,
    line width=0.65pt
  },
  tick style={
    draw=TCQCharcoal,
    line width=0.65pt
  },
  tick label style={
    font=\TCQSmall,
    text=TCQCharcoal
  },
  label style={
    font=\TCQSmall,
    text=TCQCharcoal
  },
  axis x line*=bottom,
  axis y line*=left,
  clip=false
]

\nextgroupplot[
  xmin=-0.16,
  xmax=2.35,
  ymin=33,
  ymax=63,
  xtick={0,1,2},
  xticklabels={
    {2--10},
    {11--30},
    {31+}
  },
  ytick={35,40,45,50,55,60},
  yticklabel={\pgfmathprintnumber{\tick}\%},
  major tick length=1.5pt,
  xlabel={OOS horizon (s)},
  ylabel={Accuracy (\%)},
  ymajorgrids,
  grid style={
    draw=TCQGrid,
    line width=0.45pt
  }
]

\addplot[
  name path=baselinepath,
  draw=none
] table[
  x=x,
  y=baseline
] {data/temporal_cue_qwen.dat};

\addplot[
  name path=cuepath,
  draw=none
] table[
  x=x,
  y=cue
] {data/temporal_cue_qwen.dat};

\addplot[
  draw=none,
  fill=TCQTealLight
] fill between[
  of=cuepath and baselinepath
];

\addplot+[
  color=TCQMuted,
  densely dotted,
  line width=0.9pt,
  mark=o,
  mark size=2.3pt,
  mark options={
    fill=white,
    line width=0.75pt
  },
  error bars/.cd,
    y dir=both,
    y explicit,
    error mark options={
      color=TCQMuted,
      mark size=1.4pt,
      line width=0.6pt
    },
    error bar style={
      line width=0.75pt
    }
] table[
  x=x,
  y=baseline,
  y error minus=bminus,
  y error plus=bplus
] {data/temporal_cue_qwen.dat};

\addplot+[
  color=TCQQwenBlue,
  solid,
  line width=1pt,
  mark=square*,
  mark size=2.2pt,
  mark options={
    solid,
    draw=TCQQwenBlue,
    fill=TCQQwenBlue
  },
  error bars/.cd,
    y dir=both,
    y explicit,
    error mark options={
      color=TCQQwenBlue,
      mark size=1.4pt,
      line width=0.6pt
    },
    error bar style={
      line width=0.75pt
    }
] table[
  x=x,
  y=cue,
  y error minus=cminus,
  y error plus=cplus
] {data/temporal_cue_qwen.dat};

\nextgroupplot[
  xmin=-1.5,
  xmax=12.5,
  ymin=-0.55,
  ymax=5.55,
  xtick={0,4,8,12},
  xticklabels={0,4,8,12},
  ytick={5,4,3,2,1,0},
  yticklabels={
    {Visibility},
    {Nearest},
    {O--C dir.},
    {O--C dist.},
    {O--O dir.},
    {O--O dist.}
  },
  xlabel={Temporal cue gain wrt baseline (\%)},
  xmajorgrids,
  grid style={
    draw=TCQGrid,
    line width=0.45pt
  }
]

\path[
  fill=TCQNegativeArea,
  draw=none
] (axis cs:-1.5,-0.55)
  rectangle
  (axis cs:0,5.55);

\draw[
  TCQMuted,
  densely dashed,
  line width=0.55pt
] (axis cs:0,-0.55) -- (axis cs:0,5.55);

\draw[
  TCQLight,
  line width=0.55pt
] (axis cs:-1.5,4.5) -- (axis cs:12.5,4.5);

\draw[
  TCQLight,
  line width=0.55pt
] (axis cs:-1.5,3.5) -- (axis cs:12.5,3.5);

\TCQQuestionPoint{TCQTeal}{7.5}{2.7}{2.7}{5}
\TCQQuestionPoint{TCQMuted}{3.2}{2.0}{2.0}{4}
\TCQQuestionPoint{TCQQFiveA}{8.4}{2.6}{2.6}{3}
\TCQQuestionPoint{TCQQFiveB}{2.5}{1.2}{1.2}{2}
\TCQQuestionPoint{TCQQFiveC}{1.9}{2.7}{2.7}{1}
\TCQQuestionPoint{TCQQFiveD}{4.7}{1.7}{1.7}{0}

\end{groupplot}

\path
  (group c1r1.north west) --
  coordinate[midway] (tcqlegendcenter)
  (group c2r1.north east);

\draw[
  TCQMuted,
  densely dotted,
  line width=0.9pt
]
  ([xshift=-58pt,yshift=9pt]tcqlegendcenter) --
  node[
    midway,
    circle,
    draw=TCQMuted,
    fill=white,
    inner sep=1pt
  ] {}
  ([xshift=-44pt,yshift=9pt]tcqlegendcenter);

\node[
  anchor=west,
  font=\TCQSmall,
  text=TCQCharcoal
] at ([xshift=-40pt,yshift=9pt]tcqlegendcenter)
{Baseline};

\draw[
  draw=TCQQwenBlue,
  line width=1pt
]
  ([xshift=8pt,yshift=9pt]tcqlegendcenter) --
  ([xshift=22pt,yshift=9pt]tcqlegendcenter);

\node[
  inner sep=0pt
] at ([xshift=15pt,yshift=9pt]tcqlegendcenter)
{\textcolor{TCQQwenBlue}{\rule{4pt}{4pt}}};

\node[
  anchor=west,
  font=\TCQSmall,
  text=TCQCharcoal
] at ([xshift=26pt,yshift=9pt]tcqlegendcenter)
{Temporal cue};

\end{tikzpicture}

\caption{
\textbf{Effect of temporal cues on Qwen-3.6-27B.}
\textbf{Left:} Macro accuracy across out-of-sight (OOS) horizons, with shading indicating the improvement over the baseline.
\textbf{Right:} Paired improvement for visibility, nearest-fixture, object--camera (O--C), and object--object (O--O) direction and distance questions.
Error bars show 95\% bootstrapped confidence intervals.
}
\vspace{-15pt}
\label{fig:temporal-cue}
\end{figure}

\begin{figure*}[t]
    \centering
    \includegraphics[width=\textwidth]{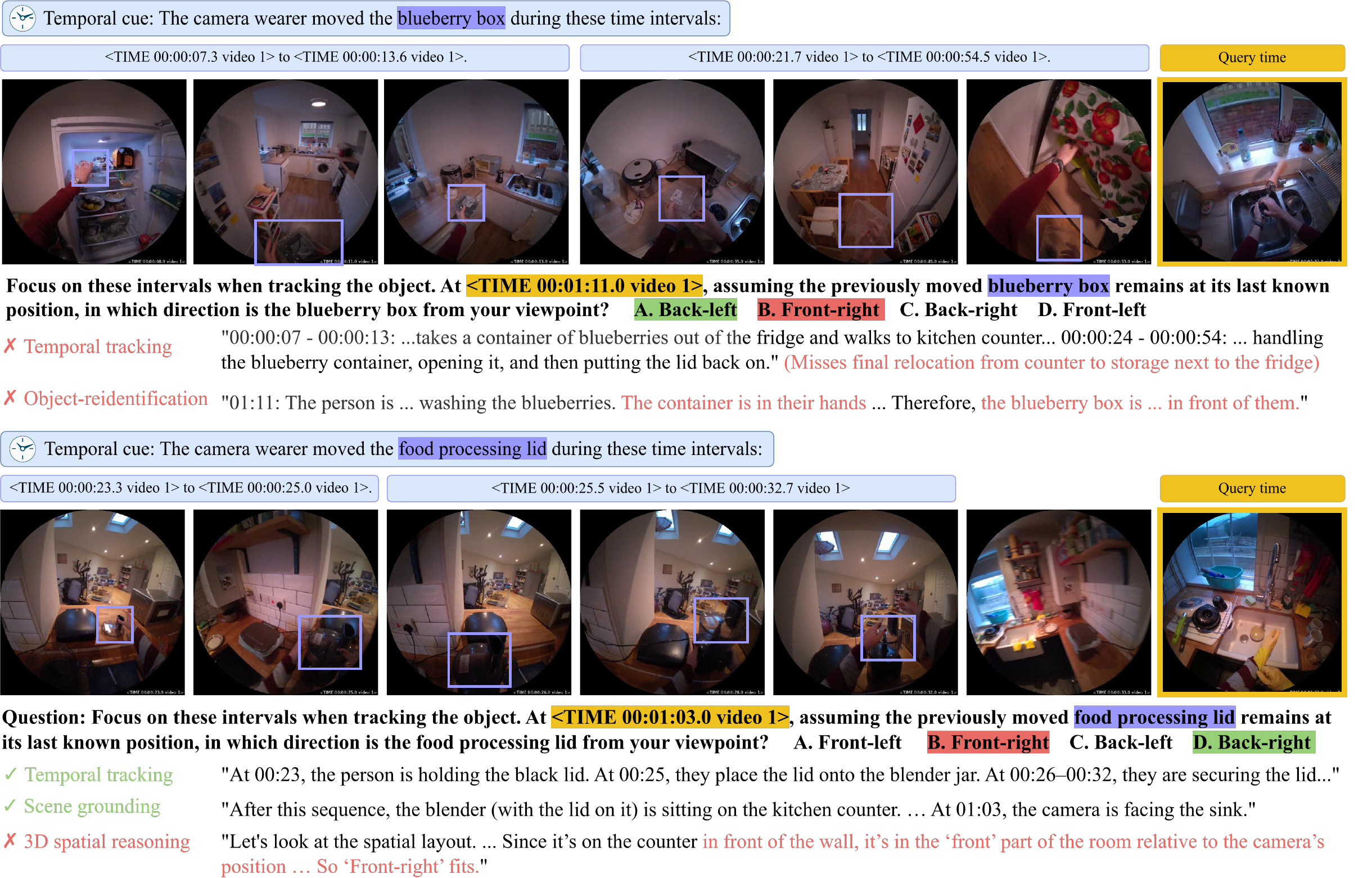}
    \caption{
    \textbf{Qualitative analysis.}
For each example, we show the provided temporal cues (blue), keyframes and Q\&A, and reasoning traces. Correct and incorrect steps are marked in green and red.
\textbf{Top:} the model misses the final relocation of the \textit{blueberry box} and misidentifies it.
\textbf{Bottom:} the model correctly tracks and grounds the \textit{food processing lid} but fails in camera-relative 3D reasoning.}
\vspace{-15pt}
    \label{fig:qual_analysis}
\end{figure*}

\subsection{Ablation Study}
\label{sec:error_analysis}
The preceding analyses suggest two upstream sources of error, namely locating the state-changing interactions and retaining the resulting state, and accounting for the target's visibility status at query time. We test these hypotheses on Qwen-3.6-27B, the strongest model in our main evaluation.

\par\noindent
\textbf{Temporal evidence retrieval is an upstream bottleneck.}
We first prepend a textual cue that lists all intervals in which the target object was moved, e.g., ``The camera wearer moved the \textit{blueberry box} during these time intervals: \texttt{<TIME 00:00:07.3 video 1>} to \texttt{<TIME 00:00:13.6 video 1>}, \ldots''. The cue directs the model toward the state-changing interactions that determine the object's latest location. As shown in Fig.~\ref{fig:temporal-cue}, this intervention improves all downstream question types across out-of-sight horizons, supporting temporal evidence retrieval as an important upstream bottleneck.

\par\noindent
\textbf{Visibility information further improves scene grounding.}
We next add an explicit textual statement that the target is not visible at query time, on top of the temporal cue. \textit{Nearest Fixture} accuracy improves by 8\%, but 3D spatial tasks gain only 0--2.4\%. This suggests that visibility awareness helps recover the target's scene location, but leaves most geometric errors unresolved.

\subsection{Qualitative Analysis}
\label{sec:qualitative_analysis}

To understand the errors remaining after simplifying temporal retrieval, we examine Qwen-3.6-27B's reasoning traces under temporal cues with thinking enabled. Fig.~\ref{fig:qual_analysis} illustrates where failures arise along the reasoning chain. Three recurring patterns emerge:
(i) \textbf{Fine-grained evidence can still be missed.}
For a \textit{blueberry box}, the model summarizes both provided movement intervals but misses the final relocation from the counter to a storage area next to the fridge. It therefore carries an outdated location forward to the query;
(ii) \textbf{Tracking errors can corrupt the remembered state.}
In the same example, the model conflates the queried box with the blueberries being washed and concludes that the container is in the person's hands. The relevant interaction is present, but the target identity is not maintained over time; (iii) \textbf{Spatial reasoning remains a downstream bottleneck.}
For a \textit{food processing lid}, the model correctly follows the movements and recovers that the lid was left on the kitchen counter. It nevertheless maps this location incorrectly into the current camera viewpoint, predicting \textit{front-right} instead of the correct \textit{back-right}.

\section{Conclusions}

We introduced a new VQA benchmark to evaluate out-of-sight spatiotemporal reasoning in dynamic egocentric videos, addressing a key gap in literature. By querying relocated objects only after they are unobservable, and decomposing the task into eight questions and four capabilities, \textsc{Beyond3D} tests whether VLMs can update an object's spatial state, retain it beyond visibility, and reason from it in space and time. Experiments show that recent VLMs remain far from reliable, with errors compounding across retrieving relevant past events, maintaining object state over time, and reasoning from that state once the object is no longer visible. Future work should therefore move beyond current-view perception toward models that explicitly update and preserve persistent latent representations of dynamic scenes over time. We believe \textsc{Beyond3D} provides a useful benchmark for measuring progress toward this capability in egocentric video understanding.

\paragraph{Acknowledgements}
We thank Xiaoxuan Cheng for assistance with executing experiments on the cluster.

{
    \small
    \bibliographystyle{ieeenat_fullname}
    \bibliography{main}
}

\clearpage
\maketitlesupplementary

\section{Supplementary}
\label{sec:Supplementary}
\paragraph{Overview.}
In this supplementary material, we provide additional information, visualizations, and analyses that complement the main paper. 
Sec.~\ref{sec:supplementary_Benchmark_and_Statistics} expands on the benchmark construction, including the question templates, answer and distractor generation, and dataset distributions. 
Sec.~\ref{sec:supplementary_Technical_Details_Model_Inference} provides implementation details for model inference, including visual preprocessing and the prompts used for the baseline and oracle interventions. 
Sec.~\ref{sec:Supplementary_visibility_track_construction} describes the visibility-track construction pipeline in detail, with additional visual illustrations of object-location inference and cross-view projection. 
Finally, Sec.~\ref{sec:Supplementary_validation} reports the results of the human validation of these visibility tracks. 
\section{Benchmark and Statistics}
\label{sec:supplementary_Benchmark_and_Statistics}
\begin{table*}[t]
\centering
\caption{\textbf{Question templates and answer choices.}
\texttt{[TIME]}, \texttt{[OBJECT]}, and \texttt{[REF]} denote the query
timestamp, target object, and visible reference object, respectively.}
\label{tab:question_templates}

\footnotesize
\setlength{\tabcolsep}{4pt}
\renewcommand{\arraystretch}{1.08}

\begin{tabularx}{\textwidth}{@{}p{2.25cm} X p{4.05cm}@{}}
\toprule
\textbf{Question} & \textbf{Exact template} & \textbf{Answer choices} \\
\midrule

\multicolumn{3}{@{}l}{\textbf{\emph{Visual Grounding}}} \\[-2pt]

Visibility Check &
At \texttt{[TIME]}, is the previously moved \texttt{[OBJECT]} visible in the current frame?
&
\emph{No}; \emph{Yes}
\\

\addlinespace[2pt]
\multicolumn{3}{@{}l}{\textbf{\emph{Temporal Grounding}}} \\[-2pt]

Last Visible Time &
Which timestamp is closest to when the \texttt{[OBJECT]} was last visible?
&
5 timestamps:
\texttt{HH:MM:SS --- N seconds before the end}
\\

Last Placement Time &
The \texttt{[OBJECT]} was moved earlier in the video. Which timestamp is closest to when it last stopped being moved?
&
5 timestamps:
\texttt{HH:MM:SS --- N seconds before the end}
\\

\addlinespace[2pt]
\multicolumn{3}{@{}l}{\textbf{\emph{Scene Localization}}} \\[-2pt]

Nearest Fixture \emph{(non-counter variant)} &
At \texttt{[TIME]}, based on the last known position of the \texttt{[OBJECT]} that was moved earlier, which fixture type is closest to it?
&
5 fixture types
\\

\addlinespace[2pt]

Nearest Fixture \emph{(counter variant)} &
At \texttt{[TIME]}, based on the last known position of the \texttt{[OBJECT]} that was moved earlier, which counter area is closest to it?
&
3--6 kitchen-dependent counter areas
\\

\addlinespace[2pt]
\multicolumn{3}{@{}l}{\textbf{\emph{3D Spatial Perception}}} \\[-2pt]

Object--Camera Direction &
At \texttt{[TIME]}, assuming the previously moved \texttt{[OBJECT]} remains at its last known position, in which direction is the \texttt{[OBJECT]} from your viewpoint?
&
\emph{Front-right}; \emph{Back-right};
\emph{Front-left}; \emph{Back-left}
\\

Object--Camera Distance &
At \texttt{[TIME]}, assuming the previously moved \texttt{[OBJECT]} remains at its last known position, what is the distance between the camera and where the \texttt{[OBJECT]} was left?
&
\emph{Under 1 m};
\emph{1 to under 1.5 m};
\emph{1.5 m or more}
\\

Object--Object Direction &
At \texttt{[TIME]}, assuming the previously moved \texttt{[OBJECT]} remains at its last known position, where is it relative to the \texttt{[REF]} (marked in red in the current frame) from your viewpoint?
&
\emph{12 to 4:30 o'clock};
\emph{4:30 to 7:30 o'clock};
\emph{7:30 to 12 o'clock}
\\

Object--Object Distance &
At \texttt{[TIME]}, assuming the previously moved \texttt{[OBJECT]} remains at its last known position, how far is it relative to the \texttt{[REF]} (marked in red in the current frame)?
&
\emph{Under 1 m};
\emph{1 to under 1.5 m};
\emph{1.5 m or more}
\\

\bottomrule
\end{tabularx}
\end{table*}
\subsection{Question Templates and Answer Choices}

\label{sec:QA_template}

Each question is constructed at a query time $T_q$ for a stationary and out-of-sight object that was moved earlier in the video using natural language template with placeholders. In the templates of Tab.~\ref{tab:question_templates}, \texttt{[OBJECT]} denotes the target object, \texttt{[REF]} denotes a visible reference object, and \texttt{[TIME]} denotes the query timestamp in the format
\texttt{<TIME HH:MM:SS.s video 1>}.
\subsection{Technical Details for Answer and Distractor Construction}
\label{sec:Supplementary_vqa_construction}
Each question is constructed as a multiple-choice question in two stages: (i) derive the ground-truth answer, and (ii) construct the remaining options. All multiple-choice answers are randomly shuffled before being added to the benchmark.

\medskip
\noindent\textbf{\emph{Ground-truth answer derivation}}

\smallskip
\noindent\emph{Temporal grounding.}
The correct answer is the timestamp of the relevant visibility transition or placement event, obtained from the visibility tracks and movement annotations.

\smallskip
\noindent\emph{Scene localization. }
For each placement event, the HD-EPIC~\cite{perrett2025hdepic} movement annotation provides the fixture closest to the object at the end of the movement. The fixture distribution is highly imbalanced: among $16{,}213$ placements with a known fixture, $63.2\%$ occur on counters, followed by the sink ($9.7\%$), hob ($6.7\%$), cupboard ($5.1\%$), dishwasher ($3.4\%$), and drawer ($3.3\%$). Collapsing all counters into a single \emph{counter} label would therefore introduce a strong answer bias and discard spatial detail.

The annotations contain 63 counter-surface instances across nine kitchens, represented only by non-semantic IDs such as \texttt{counter\_001}. We manually map each instance to a unique, human-readable description using nearby fixtures, e.g., \textit{``counter area between the fridge and the hob''}. We favor neutral landmarks to reduce object--location shortcuts; for example, \textit{``next to the microwave''} is preferred over \textit{``next to the sink''} when the latter may reveal the likely location of objects such as a sponge. Non-counter placements use the annotated fixture category directly.

\smallskip
\noindent\emph{3D spatial perception. }
We derive all 3D answers from the annotated world-frame object positions and the camera pose at query time $T_q$. Camera-relative questions express the target in the camera coordinate system at $T_q$, whereas object-relative questions translate the origin to a confidently visible reference object while preserving the axis orientation defined by the camera pose at $T_q$.

Let $c_q\in\mathbb{R}^3$ denote the camera center at query time $T_q$ in world coordinates, and let
$R_q\in SO(3)$ denote the corresponding world-to-camera rotation. The target's last known
world-frame position $\ell_o(T_q)$ is expressed in the camera frame as
\[
p_o^{\mathrm{cam}}
=
R_q\bigl(\ell_o(T_q)-c_q\bigr).
\]
Thus, the camera-relative direction is determined by the orientation of
$p_o^{\mathrm{cam}}$, while the camera-relative distance is
\[
d_{o,\mathrm{cam}}=\left\|p_o^{\mathrm{cam}}\right\|_2.
\]

For object-relative questions, we select a distinct reference object $r$ that is confidently
visible at $T_q$, with world-frame position $\ell_r(T_q)$. We translate the coordinate origin
from the camera center to the reference object while retaining the camera orientation:
\[
p_{o\mid r}^{\mathrm{cam}}
=
R_q\bigl(\ell_o(T_q)-\ell_r(T_q)\bigr).
\]
Equivalently, in homogeneous coordinates this transformation is
\[
T_{q,r}
=
\begin{bmatrix}
R_q & -R_q\ell_r(T_q)\\
\mathbf{0}^{\top} & 1
\end{bmatrix}.
\]
Hence, $p_{o\mid r}^{\mathrm{cam}}$ is the vector from the reference object to the remembered
target location, expressed along the camera-oriented axes at $T_q$. Its orientation determines
the object-relative direction answer, and
\[
d_{o,r}=\left\|p_{o\mid r}^{\mathrm{cam}}\right\|_2
       =\left\|\ell_o(T_q)-\ell_r(T_q)\right\|_2
\]
determines the object-relative distance. The resulting directions and distances are mapped to
the predefined categorical answer bins used by each question type.

\medskip
\noindent\textbf{\emph{Option construction}}

\smallskip
\noindent\emph{Temporal distractors.}
Each temporal question contains the correct timestamp and four hard negatives. Candidate timestamps are grouped into four bins by absolute temporal distance from the correct answer, \emph{near} ($\pm$1--2\,s), \emph{medium} ($\pm$3--4\,s), \emph{far} ($\pm$5--6\,s), and \emph{very far} ($\pm$7--30\,s), and one negative is sampled from each bin. We prioritize timestamps corresponding to other visibility or movement events. If none are available, we sample a regular timestamp from the same bin.

\smallskip
\noindent\emph{Scene-localization distractors.}
For counter placements, distractors are other counter-area descriptions from the same kitchen. For non-counter placements, they are sampled from alternative fixture categories, such as \textit{counter}, \textit{cupboard}, \textit{dishwasher}, or \textit{drawer}.

\textit{Nearest Fixture} is therefore the only question type whose number of options varies. Non-counter placements always yield five options, whereas counter placements yield between three and six, since a kitchen with few annotated counter areas admits fewer plausible alternatives. The chance level reported for this question type in the main results table is consequently not $1/k$ for a fixed $k$, but the mean of $1/k_i$ over the $1{,}000$ questions, which evaluates to $22.7\%$. All other question types have a fixed option count, giving the $50.0\%$, $20.0\%$, $25.0\%$, and $33.3\%$ chance levels of the remaining columns.

\smallskip
\noindent\emph{3D spatial options.}
Direction and distance are discretized into predefined, exhaustive categories, which directly define the answer choices.

\subsection{Answer Distribution}


\definecolor{posA}{HTML}{5F86BB}
\definecolor{posB}{HTML}{83A1C9}
\definecolor{posC}{HTML}{A8BDD8}
\definecolor{posD}{HTML}{CAD6E7}
\definecolor{camdistA}{HTML}{91BC8C}
\definecolor{camdistB}{HTML}{ACCCAA}
\definecolor{camdistC}{HTML}{D0E0CE}
\definecolor{objposA}{HTML}{9889BF}
\definecolor{objposB}{HTML}{B2A6CF}
\definecolor{objposC}{HTML}{D0C9E1}
\definecolor{objdistA}{HTML}{E2A354}
\definecolor{objdistB}{HTML}{EBC083}
\definecolor{objdistC}{HTML}{F3D9B5}
\definecolor{chronoTwo}{HTML}{4C78A8}
\definecolor{chronoThree}{HTML}{D07A6A}
\definecolor{distgrid}{HTML}{ECECEC}
\definecolor{disttext}{HTML}{303030}

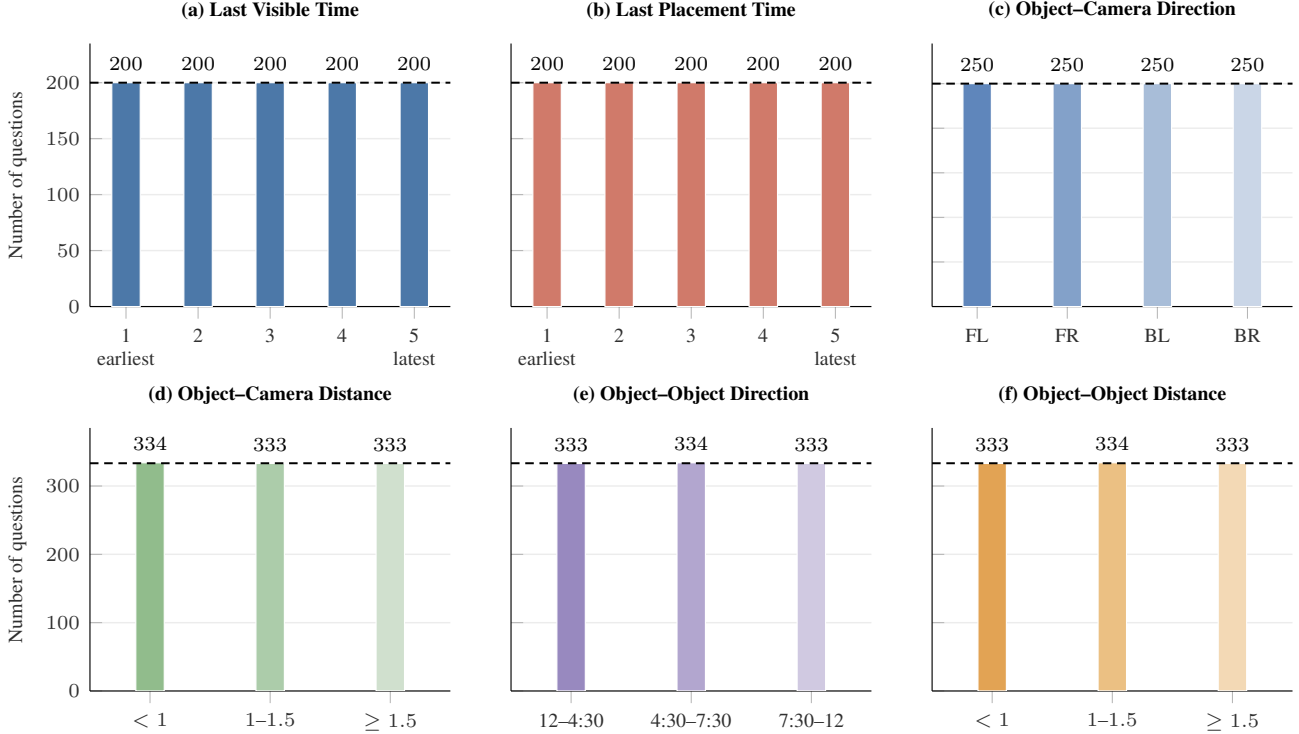
\begin{figure*}[t]
\centering
\begin{adjustbox}{width=\textwidth}
\begin{tikzpicture}
\begin{groupplot}[
    group style={
        group size=3 by 2,
        horizontal sep=0.75cm,
        vertical sep=1.5cm
    },
    width=0.255\textwidth,
    height=3.25cm,
    scale only axis,
    ybar,
    xmin=0.5,
    ymin=0,
    ymajorgrids,
    grid style={draw=distgrid,line width=0.5pt},
    axis x line*=bottom,
    axis y line*=left,
    tick style={draw=gray!60},
    tick label style={font=\scriptsize,color=disttext},
    xticklabel style={align=center},
    label style={font=\scriptsize,color=disttext},
    title style={font=\scriptsize\bfseries,yshift=-1pt},
    nodes near coords,
    point meta=y,
    every node near coord/.append style={font=\scriptsize,yshift=1pt}
]

\nextgroupplot[
    title={(a) Last Visible Time},
    xmax=5.5,
    xtick={1,2,3,4,5},
    xticklabels={1\\earliest,2,3,4,5\\latest},
    ymax=235,
    ytick={0,50,100,150,200},
    ylabel={Number of questions}
]
\addplot[fill=chronoTwo,draw=white]
coordinates {(1,200) (2,200) (3,200) (4,200) (5,200)};
\addplot[black,densely dashed,line width=0.7pt,sharp plot,forget plot,nodes near coords={}]
coordinates {(0.5,200) (5.5,200)};

\nextgroupplot[
    title={(b) Last Placement Time},
    xmax=5.5,
    xtick={1,2,3,4,5},
    xticklabels={1\\earliest,2,3,4,5\\latest},
    ymax=235,
    ytick={0,50,100,150,200},
    yticklabels={}
]
\addplot[fill=chronoThree,draw=white]
coordinates {(1,200) (2,200) (3,200) (4,200) (5,200)};
\addplot[black,densely dashed,line width=0.7pt,sharp plot,forget plot,nodes near coords={}]
coordinates {(0.5,200) (5.5,200)};

\nextgroupplot[
    title={(c) Object--Camera Direction},,
    xmax=4.5,
    xtick={1,2,3,4},
    xticklabels={FL,FR,BL,BR},
    ymax=295,
    ytick={0,50,100,150,200,250},
    yticklabels={}
]
\addplot[fill=posA,draw=white,bar shift=0pt] coordinates {(1,250)};
\addplot[fill=posB,draw=white,bar shift=0pt] coordinates {(2,250)};
\addplot[fill=posC,draw=white,bar shift=0pt] coordinates {(3,250)};
\addplot[fill=posD,draw=white,bar shift=0pt] coordinates {(4,250)};
\addplot[black,densely dashed,line width=0.7pt,sharp plot,forget plot,nodes near coords={}]
coordinates {(0.5,250) (4.5,250)};

\nextgroupplot[
    title={(d) Object--Camera Distance},
    xmax=3.5,
    xtick={1,2,3},
    xticklabels={$<1$,$1$--$1.5$,$\geq1.5$},
    ymax=385,
    ytick={0,100,200,300},
    ylabel={Number of questions}
]
\addplot[fill=camdistA,draw=white,bar shift=0pt] coordinates {(1,334)};
\addplot[fill=camdistB,draw=white,bar shift=0pt] coordinates {(2,333)};
\addplot[fill=camdistC,draw=white,bar shift=0pt] coordinates {(3,333)};
\addplot[black,densely dashed,line width=0.7pt,sharp plot,forget plot,nodes near coords={}]
coordinates {(0.5,333.333) (3.5,333.333)};

\nextgroupplot[
    title={(e) Object--Object Direction},
    xmax=3.5,
    xtick={1,2,3},
    xticklabels={12--4:30,4:30--7:30,7:30--12},
    ymax=385,
    ytick={0,100,200,300},
    yticklabels={}
]
\addplot[fill=objposA,draw=white,bar shift=0pt] coordinates {(1,333)};
\addplot[fill=objposB,draw=white,bar shift=0pt] coordinates {(2,334)};
\addplot[fill=objposC,draw=white,bar shift=0pt] coordinates {(3,333)};
\addplot[black,densely dashed,line width=0.7pt,sharp plot,forget plot,nodes near coords={}]
coordinates {(0.5,333.333) (3.5,333.333)};

\nextgroupplot[
    title={(f) Object--Object Distance},
    xmax=3.5,
    xtick={1,2,3},
    xticklabels={$<1$,$1$--$1.5$,$\geq1.5$},
    ymax=385,
    ytick={0,100,200,300},
    yticklabels={}
]
\addplot[fill=objdistA,draw=white,bar shift=0pt] coordinates {(1,333)};
\addplot[fill=objdistB,draw=white,bar shift=0pt] coordinates {(2,334)};
\addplot[fill=objdistC,draw=white,bar shift=0pt] coordinates {(3,333)};
\addplot[black,densely dashed,line width=0.7pt,sharp plot,forget plot,nodes near coords={}]
coordinates {(0.5,333.333) (3.5,333.333)};

\end{groupplot}
\end{tikzpicture}
\end{adjustbox}

\caption{\textbf{Controlled correct-answer distributions.}
\textbf{(a--b)} Chronological answer-rank distributions for \textit{Last Visible Time} and \textit{Last Placement Time}.
Chronological rank is obtained by sorting the five displayed timestamp choices from earliest to latest; rank 1 is the earliest and rank 5 the latest. Each rank is correct for exactly 200 of 1,000 questions in each step.
\textbf{(c--f)} Semantic correct-answer distributions for the four 3D spatial perception questions, grouped by answer meaning rather than the shuffled multiple-choice option letter. Panel (c) has four classes and is exactly balanced; panels (d--f) have three classes and are balanced to $334/333/333$. FL/FR/BL/BR denote front/back left/right, and distance values are in metres. Dashed lines mark equal-share reference counts.}
\label{fig:supp_balanced_answer_distributions}
\end{figure*}


\definecolor{fixtureBar}{HTML}{4E9485}
\definecolor{distgrid}{HTML}{ECECEC}
\definecolor{disttext}{HTML}{303030}

\begin{figure*}[t]
\centering

\begin{adjustbox}{max width=\textwidth,max totalheight=0.70\textheight}
\begin{tikzpicture}
\begin{axis}[
    width=0.78\textwidth,
    height=0.68\textheight,
    scale only axis,
    xbar,
    bar width=4.2pt,
    xmin=0,
    xmax=108,
    xtick={0,20,40,60,80,100},
    ymin=0.5,
    ymax=47.5,
    ytick={1,...,47},
    y dir=reverse,
    yticklabels={
        sink,
        counter area between the sink and the hob,
        hob,
        counter area close to the microwave,
        cupboard,
        counter area next to the window,
        fridge,
        counter area beside the hob and close to the door,
        counter area between the hob and the sink,
        counter area next to the wooden rolling cart,
        counter area beside the hob and near the sink,
        counter area next to the microwave,
        counter area under the long straight cupboard row,
        counter area above the washing machine,
        counter area between the microwave and the hob,
        counter area opposite the table,
        counter area under the L-shaped cupboard cluster,
        shelf,
        dishwasher,
        counter area above the refrigerator,
        table,
        counter area above a stack of 4 drawers,
        bin,
        drawer,
        oven,
        counter area between the fridge and the hob,
        fridge--freezer,
        counter area below the boiler,
        counter area next to the oven,
        counter area in the corner under the cupboard,
        counter area between the door and the toaster,
        counter area close to the washing machine,
        microwave,
        freezer,
        counter area next to the tall floor-standing cupboard,
        storage,
        counter area opposite the refrigerator,
        floor,
        washing machine,
        counter area in the corner next to the sink,
        top cupboard,
        top microwave,
        windowsill,
        hook,
        top fridge,
        counter area above the dishwasher,
        counter area next to the refrigerator
    },
    xmajorgrids,
    grid style={draw=distgrid,line width=0.5pt},
    axis x line*=bottom,
    axis y line*=left,
    tick style={draw=gray!60},
    tick label style={font=\scriptsize,color=disttext},
    yticklabel style={font=\scriptsize,text width=5.7cm,align=right},
    label style={font=\scriptsize,color=disttext},
    xlabel={Number of questions},
    nodes near coords,
    point meta=x,
    every node near coord/.append style={font=\scriptsize,xshift=1pt,anchor=west}
]
\addplot[fill=fixtureBar,draw=white]
table[x=count,y=row]
{data/answer_dist_fixture.dat};
\end{axis}
\end{tikzpicture}
\end{adjustbox}

\caption{\textbf{Correct-answer distribution for \textit{Nearest Fixture}.}
\textit{Nearest Fixture} asks for the fixture type or counter area closest to the target object's last known
position. Bars show the number of questions for each correct label. Unlike the other question types, this
distribution is not balanced but retains the naturally occurring fixture frequencies. All 47 labels occurring
across the nine kitchens are shown, and the counts sum to the $1{,}000$ out-of-sight anchors.}
\label{fig:supp_step4_distribution}
\end{figure*}
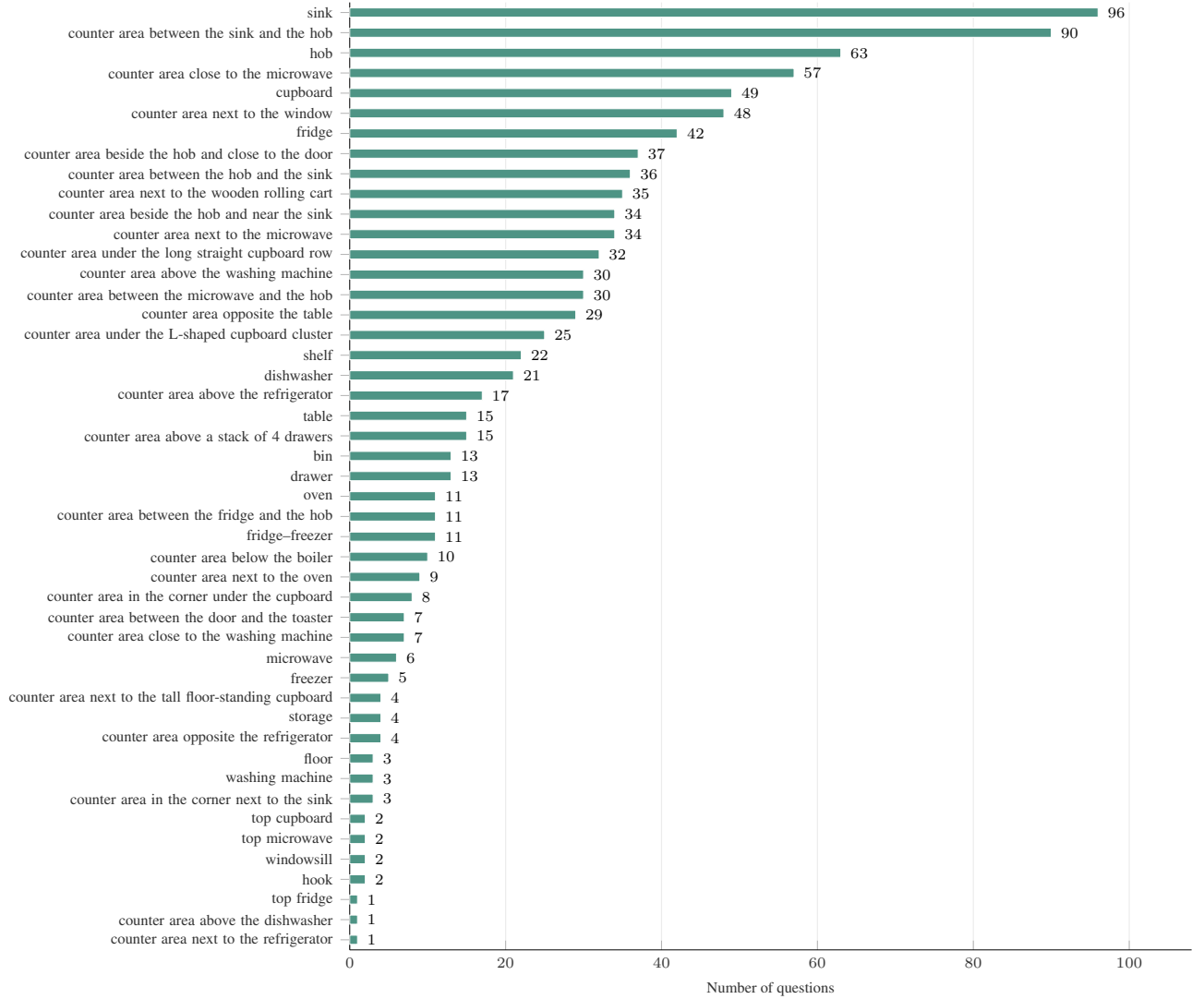

To reduce the possibility of exploiting answer-frequency shortcuts, we control the correct-answer distributions for the temporal and spatial questions. For \textit{Last Visible Time} and \textit{Last Placement Time}, each question contains five timestamp choices. We sort these choices chronologically and balance which temporal position contains the correct answer: the correct timestamp is the earliest choice for 200 questions, the second earliest for 200, and so on up to the latest choice. Thus, each of the five chronological ranks occurs equally often as the correct answer. The 3D spatial perception questions are balanced as evenly as the sample size allows across their semantic answer classes. \textit{Object--Camera Direction} has four classes and is exactly balanced at $250$ questions each, while \textit{Object--Camera Distance}, \textit{Object--Object Direction}, and \textit{Object--Object Distance} have three classes and are balanced at $334/333/333$. \textit{Nearest Fixture} is treated differently since its answers retain the naturally occurring fixture-label frequencies rather than being artificially balanced. The resulting distributions are shown in Supplementary Figs.~\ref{fig:supp_balanced_answer_distributions} and~\ref{fig:supp_step4_distribution}.
\section{Technical Details for Model Inference}
\label{sec:supplementary_Technical_Details_Model_Inference}
\subsection{Visual Input Preprocessing}
\label{sec:Visual_input_preprocessing}
 All videos are temporally sampled at 1\,fps, resized to $448 \times 448$ pixels. Pixels outside the circular fisheye field of view are masked in black. Each sampled frame is annotated near the bottom-right corner, outside the visible camera field, with a timestamp token of the form \texttt{<TIME HH:MM:SS.s video 1>}. For the \textit{Object--Object Direction} and \textit{Object--Object Distance} questions, the model must reason about the spatial relationship between the out-of-sight target object $o$ and a reference object $r$ that remains visible at the query timestamp. 
 
\begin{figure}[t]
\centering
\includegraphics[width=\columnwidth]{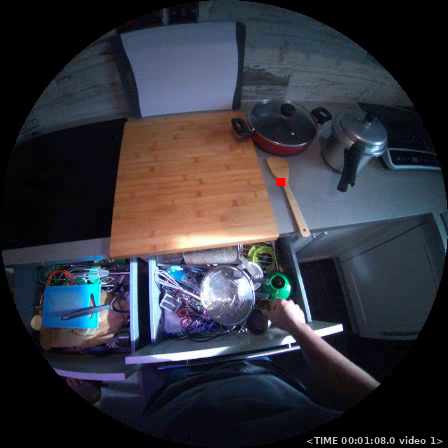}
\caption{\textbf{Example preprocessed query frame.}
The timestamp is placed in the masked region outside the fisheye field of view. For object-relative spatial questions, the visible reference object is indicated by a red marker.}
\label{fig:input}
\end{figure}

 To unambiguously identify $r$, we follow prior work on visual prompting and egocentric spatial reasoning that uses overlaid markers to indicate the queried object~\cite{shtedritski2023clip,yang2023setofmark,ravi2025outofsight}. We overlay an $8 \times 8$ red marker at its projected image location in the query-time frame only. The marker is chosen to be clearly visible after resizing while minimally occluding the surrounding visual content; restricting it to the query-time frame avoids providing additional information about the reference object's trajectory. An example of the processed frame is shown in Fig.~\ref{fig:input}.
 
 For each evaluation sample, the model receives the video prefix from $t=0$ to the query time $T_q$, inclusive. Models that can accommodate the full prefix receive all sampled frames, while for models with shorter context limits, frames are further uniformly subsampled to fit the available context. The frame at $T_q$ is always retained to preserve the visual state at the query time.

\subsection{Inference Prompt}
\label{sec:Inference_prompt}
\paragraph{Baseline setup.}
Each sample consists of a fixed system prompt, the input video, and a multiple-choice question with its answer options. We use the same system prompt for all models:
{\small\itshape
``You are a helpful assistant trained to answer spatial and visual questions based on egocentric videos. Use the video to answer the question. The video is sampled at 1 frame per second.''}

The question and answer choices are provided using the following prompt:

{\small\itshape
``Question: [QUESTION]

Options:

A. [OPTION A]

B. [OPTION B]

\ldots

Select the best option and output only its letter.''}

\paragraph{Temporal-cue intervention.}
For the temporal-cue experiments, we prepend the annotated intervals during which the target object was moved:
{\small\itshape
``Temporal cue: The camera wearer moved the [OBJECT] during these time intervals: [START$_1$] to [END$_1$]; \ldots; [START$_n$] to [END$_n$]. Focus on these intervals when tracking the object.''}

For example:
{\small\itshape
``Temporal cue: The camera wearer moved the fork during these time intervals: \texttt{<TIME 00:00:03.0 video 1>} to \texttt{<TIME 00:00:14.3 video 1>}; \texttt{<TIME 00:00:15.5 video 1>} to \texttt{<TIME 00:00:26.4 video 1>}. Focus on these intervals when tracking the object.''}
\paragraph{Visibility-cue intervention.}
For the visibility-cue experiments, we additionally state that the target is not visible at query time. Specifically, downstream questions are rewritten into the following form, where \texttt{[QUESTION BODY]} is the template of Tab.~\ref{tab:question_templates} with its leading ``At \texttt{[TIME]}, assuming the previously moved \texttt{[OBJECT]} remains at its last known position,'' clause removed, so that the clause is not stated twice:
{\small\itshape
``At [QUERY TIME], the previously moved [OBJECT] is no longer visible. Assuming it remains at its last known position, [QUESTION BODY]''}

For example:
{\small\itshape
``At \texttt{<TIME 00:01:00.0 video 1>}, the previously moved cloth is no longer visible. Assuming it remains at its last known position, in which direction is the cloth from your viewpoint?''}

\section{Technical Details for Visibility Track Construction}
\label{sec:Supplementary_visibility_track_construction}
This section provides the implementation details for the visibility track construction procedure. Tracks are sampled at 1\,fps and contain both the inferred object location and its visibility state.

\subsection{Inferring Object Locations from Movement Annotations}
\begin{figure}[t]
\centering
\includegraphics[width=\columnwidth]{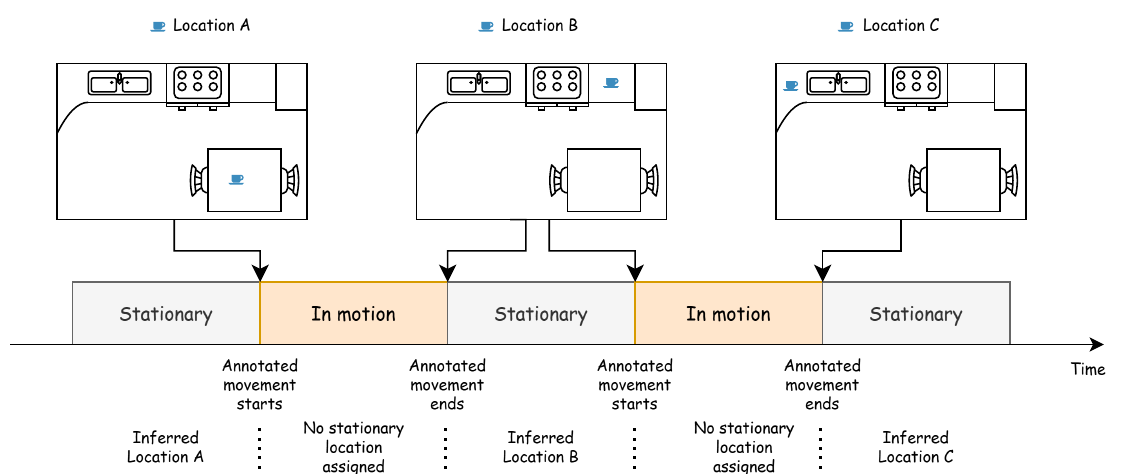}
\caption{\textbf{Object location inference from HD-EPIC annotations.}
The object is held at its first annotated start location before the first movement, marked \texttt{in\_motion} during a movement, and held at the endpoint of its most recent completed movement thereafter.}
\label{fig:loc_inf}
\end{figure}

HD-EPIC~\cite{perrett2025hdepic} annotates each object movement with its temporal interval and observations at the beginning and end of the movement. These observations include a 2D bounding box, segmentation mask, 3D object center, and associated scene fixture. Because intermediate object trajectories are not annotated, we infer the stationary object location using the following rules, illustrated in Fig.~\ref{fig:loc_inf}:
\begin{itemize}
\item \textbf{Before the first annotated movement:}
The object is assumed to remain at the location recorded at the beginning of its first movement.

\item \textbf{During an annotated movement:}
The object is assigned \texttt{in\_motion}. No stationary location is assigned because its trajectory between the annotated endpoints is unknown, so the three-stage procedure below is not applied to these samples. We nevertheless treat the object as visible while it is being relocated ($v_o(t)=1$), since it is in the camera wearer's hands. Because they carry no stable location, they can never themselves be query anchors, and they are excluded from the audit in Sec.~\ref{sec:Supplementary_validation}.

\item \textbf{Between movements and after the final movement:}
The object is assumed to remain at the endpoint location of its most recently completed movement.

\end{itemize}

\subsection{Determining Visibility}
\label{sec:Supplementary_determining_visibility}
For each one-second sample at which the object has an inferred stationary location, we determine its visibility using a three-stage procedure:

\begin{enumerate}
\item \textbf{Field-of-view projection:} We determine whether the object's inferred location falls within the current camera view.
\item \textbf{Fixture-aware geometric occlusion:} We check whether a scene fixture blocks the camera's view of the object.
\item \textbf{Detection-based visual confirmation:} We inspect the corresponding video frame to confirm whether the object is visible at the expected location using an open-vocabulary detection method.
\end{enumerate}

Stages~1 and~2 act on individual samples. Consecutive samples sharing the same stage-1/stage-2 outcome are then merged into a continuous \emph{interval}, and Stage~3 operates at interval granularity, where it scores each candidate interval by the fraction of its frames in which the detector confirms the object, and assigns the resulting state to the interval as a whole. Every sample therefore ends up with exactly one state, and every interval is state-homogeneous by construction.

\paragraph{Stage 1: Field-of-view projection.}
For stationary objects, projecting only the annotated 3D centroid is unreliable near image boundaries, as objects may remain partially visible even if their centroid is off-screen. To account for this, we determine if an object is within the field of the camera view using the following procedure:

\begin{itemize}
    \item \textbf{Approximating spatial extent:} We represent the object using nine points from its annotated endpoint bounding box: the center, four corners, and four edge midpoints. These points are back-projected onto a plane that passes through the object's 3D centroid and is parallel to the camera's image plane, forming a stable 3D footprint in world coordinates that is reused while the object remains stationary. 
    
    \item \textbf{Camera projection and masking:} At each sampled time point, we reproject the nine points into the current camera view using the corresponding camera pose provided by the HD-EPIC annotations and the FISHEYE624 fisheye camera model of Project Aria~\cite{engel2023projectaria}. We use devignetted RGB frames, in which lens-vignetting effects are corrected and valid image content is limited to the circular region within the frame. The same frames and valid-image mask are used in later detection stages and during benchmark evaluation to ensure consistency. A projected point is considered invalid if it lies behind the camera or outside this region. 
    \item \textbf{State assignment:} The object is assigned \texttt{in\_view} if more than 50\% of the points (i.e. at least five of the nine support points) are valid; otherwise, it is assigned the state \texttt{out\_of\_view}.
\end{itemize}

\paragraph{Stage 2: Fixture-aware geometric occlusion.}
For each in-view object, we cast rays from the camera center toward all nine support points and intersect them with the kitchen mesh. A ray is considered blocked if its first intersection lies at least $\delta=10\,\mathrm{cm}$ closer to the camera than the corresponding target support point. We then calculate the fraction of rays that are blocked. If at least $50\%$ of the rays are blocked, the object is considered geometrically occluded.

The static digital twin does not track the real-time state of movable parts, such as open refrigerator doors or extended drawers. Consequently, an object placed inside an open cabinet might falsely appear occluded by the digital twin's default closed-door geometry. If the majority blockage is attributed to the openable fixture containing the object, we instead assign \texttt{fixture\_ambiguous} and defer the final visibility decision to the detection-based visual confirmation stage.

\paragraph{Stage 3: Detection-based visual confirmation.}
Samples that lie within the camera's field of view and are not classified as \texttt{occluded} are verified using an open-vocabulary object detector. We process the corresponding frames at 1\,fps, apply the same valid-image mask used in the previous stages, and query OWLv2~\cite{minderer2023scalingOWLv2} with the object's name. A detection is matched to the target only if its bounding box, enlarged by $20$ pixels, contains the object's projected anchor point. If multiple detections satisfy this condition, we select the one whose center is closest to the target object. This geometric constraint helps distinguish between multiple objects of the same category.

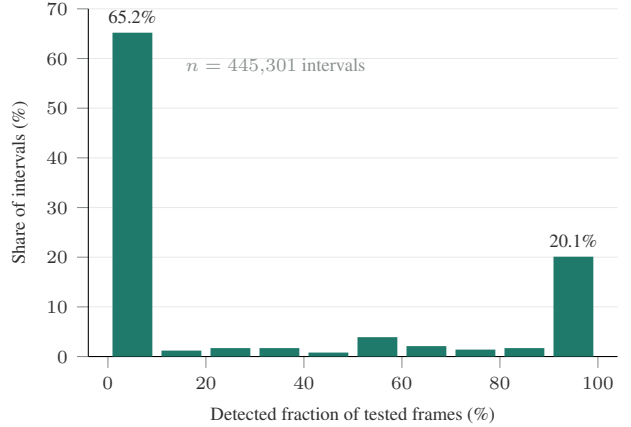
\begin{figure}[t]
\centering

\definecolor{dfbar}{HTML}{1F7A6B}
\definecolor{dfgrid}{HTML}{E7EAE7}
\definecolor{dftext}{HTML}{303030}
\definecolor{dfmuted}{HTML}{8A918C}

\begin{tikzpicture}
\begin{axis}[
    width=0.84\columnwidth, height=4.6cm, scale only axis,
    ybar, bar width=15pt, bar shift=0pt,
    xmin=-4, xmax=104,
    ymin=0,  ymax=70,
    xtick={0,20,40,60,80,100},
    ytick={0,10,20,30,40,50,60,70},
    ymajorgrids, grid style={draw=dfgrid, line width=0.4pt},
    axis x line*=bottom, axis y line*=left,
    tick align=outside, tick style={draw=dfmuted, line width=0.4pt},
    tick label style={font=\scriptsize, color=dftext},
    label style={font=\scriptsize, color=dftext},
    xlabel={Detected fraction of tested frames (\%)},
    ylabel={Share of intervals (\%)},
    clip=false,
]

\addplot[fill=dfbar, draw=none] coordinates {
    ( 5,65.2) (15, 1.2) (25, 1.7) (35, 1.7) (45, 0.8)
    (55, 3.9) (65, 2.1) (75, 1.4) (85, 1.7) (95,20.1)
};

\node[font=\scriptsize, color=dftext, anchor=south] at (axis cs: 5,65.2) {65.2\%};
\node[font=\scriptsize, color=dftext, anchor=south] at (axis cs:95,20.1) {20.1\%};

\node[anchor=north west, font=\scriptsize, color=dfmuted]
    at (axis cs:14,62) {$n = 445{,}301$ intervals};
\end{axis}
\end{tikzpicture}

\caption{\textbf{Detected-fraction distribution of scored intervals.}
Share of intervals (\%) whose detected fraction of tested frames falls in
each $10\%$ bin ($n = 445{,}301$). The distribution is strongly U-shaped:
$65.2\%$ of intervals fall in the lowest bin and $20.1\%$ in the highest,
leaving under a sixth of intervals spread across the middle.}
\label{fig:detected_fraction}
\end{figure}

An interval is assigned \texttt{detected\_visible} if the object is detected in at least half of its testable frames. This threshold is not a sensitive parameter as the detected fraction is strongly U-shaped (Fig.~\ref{fig:detected_fraction}), with $65.2\%$ of intervals in the lowest $10\%$ bin and $20.1\%$ in the highest, so only a small minority of intervals lie near the cut-off. If the object is not detected and its previous state was \texttt{fixture\_ambiguous}, the interval is assigned \texttt{occluded}. Otherwise, it is assigned the internal state \texttt{visually\_unconfirmed}. 

HD-EPIC object-association names are free-form and may contain instance indices, spelling errors, annotation notes, or descriptions that do not refer to a single visually identifiable object. We therefore manually curate the $3{,}826$ distinct raw names. Of these, $1{,}713$ are retained unchanged, $1{,}563$ are rewritten as visually groundable object names, and $550$ cannot be mapped to a single object. For each groundable name, the detector uses an ordered sequence of up to three queries, from specific to coarse; for example, \emph{air fryer drawer} $\rightarrow$ \emph{drawer} $\rightarrow$ \emph{air fryer}. A coarser query is used only if all more specific queries fail at the projected location. Ungroundable associations are assigned the \texttt{not\_groundable} state. These are not passed to the detector and are excluded from the constructed visibility track and the visibility-recall calculations. These objects will not be used for building the benchmark as well.


\section{Human Validation of the Visibility Tracks}
\label{sec:Supplementary_validation}
This section details the visibility track audit.

\paragraph{Audit protocol.}
We draw videos round-robin over participants and stratify frame times into uniform bins within each video, preferring frames that contain at least three in-view objects. The audit covers $30$ videos and all nine participants. Objects whose track is defined at the sampled timestamp are drawn as markers at their projected position. Annotators click the markers they can see, so unmarked objects are recorded as not visible, and a marker can also be flagged \emph{unsure}, which abstains from both the majority vote and the agreement statistics. The pipeline state and the detector output are hidden, so the annotator judges only the frame and the marker. Judgments are matched to the track state within a $1$\,s tolerance.
\begin{table}[t]
\centering
\caption{\textbf{Human audit of the visibility tracks.}
We compare the total number of annotated marks per visibility track state with the number of marks judged to be visible. These correspond to disagreements for the first three states, which the pipeline labels not visible, and agreements for \texttt{detected\_visible}. \textbf{Acc.} is computed as the percentage of agreeing marks. The pipeline makes no claim for \texttt{not\_groundable}. The four scored states sum to the $4{,}102$ marks used for scoring while the $4$ \texttt{in\_motion} marks among the $4{,}176$ gold marks are omitted, as the pipeline derives no visibility evidence for them.}
\label{tab:visibility_audit}
\small
\setlength{\tabcolsep}{4pt}
\renewcommand{\arraystretch}{1.08}
\begin{tabular}{@{}lrrr@{}}
\toprule
\textbf{Pipeline state} & \#\textbf{Marks} & \#\textbf{Visible} & \textbf{Acc.\ (\%)} \\
\midrule
\texttt{out\_of\_view}  & $1{,}293$ & $105$ & $91.9$ \\
\texttt{occluded} & $291$     & $11$  & $96.2$ \\
\texttt{visually\_unconfirmed}  & $1{,}510$ & $372$ & $75.4$ \\
\texttt{detected\_visible}       & $1{,}008$ & $819$ & $81.2$ \\
\midrule
\texttt{not\_groundable}     & $70$      & $-$   & $-$ \\
\bottomrule
\end{tabular}
\end{table}
Two properties of this design matter when reading the rates below. Frames are biased toward object-populated moments rather than drawn uniformly at random, and frames dense with objects are capped at $15$ markers, drawn at random among that frame's objects, so very dense frames are represented by a subset of their objects. The sampler has no knowledge of the pipeline states, so the per-state coverage in Table~\ref{tab:visibility_audit} is an outcome of this procedure rather than part of its design.

Three annotators completed $449$ frames in $4.8$ hours, of which $436$ pass the quality filter below, giving $5{,}391$ mark judgments over $344$ distinct frames. We aggregate multiply-rated marks by majority vote and discard $12$ marks that all raters called unsure and $12$ that end in a tie, leaving $4{,}176$ gold marks. Of these, $70$ are \texttt{not\_groundable} and $4$ are \texttt{in\_motion}. For neither state does the pipeline derive a visibility decision from the geometric and detection evidence that this audit tests. Therefore both cases are excluded from scoring and $4{,}102$ marks remain.

\paragraph{Quality control.}
We discard frames on which an annotator spent less than $2$\,s, since the markers cannot be inspected in that time. Because unmarked objects count as not visible, a rushed frame yields confident wrong labels rather than missing data, so filtering on dwell is necessary. This removes $13$ of $449$ frames. Median dwell per frame is $38.5$, $17.4$, and $16.9$\,s. Splitting each session into thirds, median dwell is lower in the final third than in the first for all three annotators (e.g.\ $42.9 \rightarrow 34.7$\,s), but the rate at which they call objects visible stays within $2.7$ points of its session mean for all three. Faster judgments later in a session are therefore not systematically more permissive.

\paragraph{Inter-annotator agreement.}
$50$ frames ($634$ marks) carry ratings from more than one annotator. Krippendorff's $\alpha$~\cite{krippendorff2018content} is $0.76$, the raters are unanimous on $520$ of these marks ($82.0\%$), and $12$ reach no majority, leaving $622$ multiply-rated marks in the gold set. Pairwise Cohen's $\kappa$~\cite{cohen1960coefficient} is $0.86$, $0.75$, and $0.68$, computed on the $550$, $583$, and $552$ marks that each pair rated without either annotator flagging \emph{unsure}. Part of the disagreement is a threshold effect. On the shared marks the three annotators call the object visible at rates of $34.4\%$, $31.5\%$, and $26.1\%$, and the annotator with the lowest rate is the one involved in both of the lowest pairwise $\kappa$ values.

\paragraph{Overall metrics.}
Table~\ref{tab:visibility_audit_supp} scores the pipeline against the majority human vote over the $4{,}102$ scored marks, counting every mark once. Confidence intervals come from a cluster bootstrap over whole videos ($30$ clusters, $2{,}000$ resamples) and therefore account for the correlation between marks drawn from the same video. Restricting scoring to the unanimous multiply-rated marks improves every metric, so label ambiguity makes the headline numbers conservative rather than flattering. Of the $520$ unanimous marks, $518$ fall in a state for which the pipeline makes a visibility claim and are therefore scored.

\begin{table}[t]
\centering
\caption{\textbf{Pipeline versus majority human vote.} Metrics over raw mark counts with $95\%$ cluster-bootstrap confidence intervals over videos. The last column restricts scoring to the $518$ scored marks among the $520$ multiply-rated marks on which the annotators are unanimous.}
\label{tab:visibility_audit_supp}
\small
\setlength{\tabcolsep}{4pt}
\renewcommand{\arraystretch}{1.08}
\begin{tabular}{@{}lccc@{}}
\toprule
\textbf{Metric} & \textbf{All marks} & \textbf{95\% CI} & \textbf{Unanimous} \\
\midrule
Accuracy          & $83.5$ & $[80.7,\ 85.8]$ & $87.6$ \\
Precision         & $81.2$ & $[76.9,\ 85.2]$ & $82.2$ \\
Recall            & $62.7$ & $[54.9,\ 68.5]$ & $69.3$ \\
F1                & $70.8$ & --              & $75.2$ \\
Balanced acc.     & $78.0$ & $[74.1,\ 81.0]$ & $81.9$ \\
\midrule
Audited marks     & $4{,}102$ & & $518$ \\
\bottomrule
\end{tabular}
\end{table}

\paragraph{Where the errors are.}
The $4{,}102$ scored marks contain $488$ false negatives and $189$ false positives. Of the false negatives, $372$ are \texttt{visually\_unconfirmed}, meaning the object is in view, unoccluded by static geometry, and legible to a human, but OWLv2 fails to confirm it, returning a matching box in fewer than half of the tested frames. The remaining $105$ and $11$ fall in \texttt{out\_of\_view} and \texttt{occluded}, whose labels rest on the projected footprint and the scene mesh. Occlusions caused by the object's own containing fixture are additionally required to fail a detector check before the \texttt{occluded} label is kept.

This asymmetry follows from the design. Detection can only demote an in-view, unoccluded sample to \texttt{visually\_unconfirmed} or leave a fixture-ambiguous sample \texttt{occluded}, so detector failures cost recall rather than precision. We keep \texttt{visually\_unconfirmed} separate from \texttt{out\_of\_view} and \texttt{occluded} throughout because it is the unreliable one, and query anchors are drawn only from the latter two.

\paragraph{Variation across participants.}
Balanced accuracy ranges from $66.8\%$ (P08) to $85.2\%$ (P09) in Table~\ref{tab:visibility_audit_participants}. Precision and recall vary by a comparable amount across kitchens, from $65.9$ to $93.6\%$ and from $39.4$ to $74.8\%$, which is expected given that both are set by how well the open-vocabulary detector handles that kitchen's objects. The weakest case is P08, where one of the three audited videos (P08-20240618-171546) contributes zero true positives, meaning the detector confirmed none of the objects that annotators could see there.

\begin{table}[t]
\centering
\caption{\textbf{Audit results per participant.} Percentages over raw mark counts, with \emph{Marks} the number of scored marks.}
\label{tab:visibility_audit_participants}
\small
\setlength{\tabcolsep}{5pt}
\renewcommand{\arraystretch}{1.08}
\begin{tabular}{@{}lrrrrr@{}}
\toprule
\textbf{Part.} & \textbf{Marks} & \textbf{Acc.} & \textbf{Prec.} & \textbf{Rec.} & \textbf{Bal. acc.} \\
\midrule
P01 & $474$ & $84.0$ & $77.7$ & $63.0$ & $77.8$ \\
P02 & $426$ & $81.7$ & $75.3$ & $57.5$ & $74.7$ \\
P03 & $323$ & $80.8$ & $65.9$ & $61.4$ & $74.7$ \\
P04 & $654$ & $87.9$ & $89.4$ & $69.8$ & $83.0$ \\
P05 & $204$ & $81.9$ & $93.6$ & $69.5$ & $82.2$ \\
P06 & $557$ & $82.9$ & $79.9$ & $66.8$ & $79.1$ \\
P07 & $522$ & $84.9$ & $77.2$ & $62.4$ & $77.8$ \\
P08 & $479$ & $74.7$ & $78.8$ & $39.4$ & $66.8$ \\
P09 & $463$ & $89.2$ & $88.4$ & $74.8$ & $85.2$ \\
\midrule
All & $4{,}102$ & $83.5$ & $81.2$ & $62.7$ & $78.0$ \\
\bottomrule
\end{tabular}
\end{table}

\end{document}